\documentclass[mastersthesis,12pt]{wuthesis}

\renewcommand{\thesisauthorfirstname}{Junyi}
\renewcommand{\thesisauthorlastname}{Li}

\renewcommand{\thesisdepartment}{Department of Electrical \& Systems Engineering}

\renewcommand{\thesisfield}{Engineering Data Analytics and Statistics}

\renewcommand{\thesismonth}{Fall}
\renewcommand{\thesisyear}{2026}

\renewcommand{\thesistitle}{Latent Safety Filtering for Humanoid Locomotion}

\renewcommand{\thesischair}{Hussein Sibai}

\renewcommand{\thesiscommittee}{
    Ioannis (Yiannis) Kantaros \\
    James Feher \\
    }

\usepackage{graphicx}
\usepackage{amsmath}
\usepackage{amssymb}
\usepackage{hyperref}
\usepackage{float}
\usepackage[ruled,vlined,linesnumbered]{algorithm2e}
\SetAlFnt{\normalsize}
\SetAlgoNlRelativeSize{0}
\SetAlgoInsideSkip{mediumskip}

\usepackage{xcolor}
\usepackage{hyperref}
\hypersetup{
    colorlinks,
    linkcolor={red!50!black},
    citecolor={blue!50!black},
    urlcolor={blue!80!black}
}

\begin{document}

\frontmatter
%
%


\thesistitlepage
      
\thesiscopyrightpage

\begin{singlespace}
\setcounter{page}{2}
\tableofcontents

\cleardoublepage
\phantomsection
\addcontentsline{toc}{chapter}{\listfigurename}
\listoffigures

\cleardoublepage
\phantomsection
\addcontentsline{toc}{chapter}{\listtablename}
\listoftables


\end{singlespace}


\begin{thesisacknowledgments}

I would like to express my thanks to Professor Hussein Sibai, for his help and support in the work of my Master's research. Special thanks to Professor Jim Feher for his help and support in planning my academic work, for his infinite patience and for explaining many things to me throughout my academic studies. And also thank to Professor Ioannis Kantaros for agreeing to be my advisor for this work and for his help and support.

I would also like to extend my heartfelt gratitude to my parents, for their financial and emotional support throughout my studies, enable me to pursue my academic goals and interests, and take advantage of opportunities offered to me.

I would like to thank my research collaborators Yuxuan Yang, Ihab Tabbara, Yerin Kang, and Christine Jiang for their contributions, for working with me, and for bringing different perspectives to the table. I also thank my roommate Zheng Fang for being a great roommate and for making my time at Washington University in St. Louis so enjoyable and memorable.

\end{thesisacknowledgments}


\cleardoublepage
\phantomsection

%
\begin{thesisabstract}

Safe locomotion remains a challenge for humanoid robots navigating environments with obstacles. Although learning-based locomotion controllers enable effective velocity tracking, they do not necessarily ensure navigation safety. This project develops a learning-based safety filtering framework to reduce collisions while maintaining the velocity-tracking capabilities of a nominal locomotion controller. The proposed safety filter maps visual observations and LiDAR point clouds and nominal velocity commands to safe ones. The filter is based on Hamilton-Jacobi reachability and is trained using actor-critic algorithms. The actor estimates the safest policy and the critic estimates the safety of that policy at different states. We evaluate vision-based, LiDAR-based, and multimodal safety filters, and the results demonstrate that they improve navigation safety. The LiDAR–Vision safety filters achieve the highest navigation success rate, the lowest collision rate, and the largest minimum obstacle distance, highlighting the benefits of combining complementary sensing modalities for safe navigation. We further deploy the framework on a Unitree G1 humanoid after fine-tuning the filters using trajectories collected from physical demonstrations. These findings shows the potential of learning-based safety filtering and multimodal sensing for safer humanoid locomotion.

\end{thesisabstract}


\mainmatter

\chapter{Introduction}
\label{chap:introduction}

The safe operation of autonomous robots has been a major challenge in the field of robotics. Autonomous robots are increasingly being developed to operate in complex and unstructured environments \cite{min2026autonomous,HumanoidRobotLocomotionSurveyandPerspectives}. In such environments, locomotion is a fundamental capability that enables robots to navigate their surroundings and perform tasks at different locations \cite{ha2025legged, Freyleggedrobots,KurniawanLeggedRobotLocomotion}. Many of the control policies designed for robot locomotion attempt to enable a robot to navigate through its surroundings while avoiding obstacles and traveling in a smooth and stable manner \cite{Omni-Perception,RenVB-Com}. For the highly complex and dynamical systems of the humanoid robots, safe locomotion is extremely challenging due to the high-dimensional control space. Therefore, designing locomotion controllers that can be adapted to different environments while always fulfilling given safety constraints is one of the current major topics within the robotics research community \cite{SHIELD,PengSafeBipedalRobotNav,OneFilter}.

To address these safety challenges in robot locomotion, safety filters have been widely used to apply safety constraints on unsafe nominal control. They intervene when a nominal action violates safety constraints and then adjust it to maintain safety \cite{HsuSafetyFilterReview}. 
One of the widely used safety filter methods is Control Barrier Functions (CBFs), which first define a safe set and then constrain the control to stay within this set. CBFs are efficient in terms of computation and can be even implemented in real time using quadratic programs. But for complex robotic systems, it is challenging to design valid barrier functions, especially when the system dynamics are highly nonlinear and the system has high dimensionality \cite{AmesCBF,COHENCBF,GARGCBF}. Hamilton–Jacobi (HJ) reachability is another popular safety filtering technique. It uses the system dynamics and computes a safety value function which indicates whether a state is safe or not. HJ reachability is able to handle non-linear system dynamics as well as bounded unknowns while still ensuring safety in a formal sense \cite{BansalHJReachability}. However, obtaining an accurate safety value function remains challenging, as its computation and approximation can become increasingly expensive in high-dimensional state spaces \cite{ChenHJReachability}.

Most safety filtering methods for robot control can be applied when the robot is equipped with sensors. Most safety filtering methods for robot control are based on well-defined system states and safety functions. However, acquiring safety-relevant information from sensor observations in complex environments is difficult because sensor observations do not directly map onto the robot’s physical states, which is especially true for sensors that provide high-dimensional data such as images. We need methods for latent safety filtering for control of robots in complex environments to acquire safety-relevant information from images for safe control of the robot. Moreover, different sensors provide different kinds of information about the environment \cite{TabbaraDesignLatent,OneFilter,seoUncertainty-awareLatentSF}. Therefore, the performance of safety filters that are trained with different sensors or even with combinations of different sensors makes this meaningful \cite{ZendelCamera+Lidar_fuse}. The question is how safety filters, which are trained with different sensor modalities and their combinations, can detect potential hazards and support safe locomotion.

In this thesis, we employ a learning-based Hamilton–Jacobi (HJ) reachability approach to design a safety filtering system for humanoid locomotion. Even though CBF-based methods can learn to design barrier functions from sensor data, designing such barrier functions to satisfy a given set of safety requirements is challenging, especially when dealing with high-dimensional sensors such as those used for vision. The HJ reachability formulation, on the other hand, allows us to assess the safety of a robot’s possible future states and to design its control to avoid unsafe regions of its state-space \cite{BansalHJReachability}. We therefore use the HJ reachability formulation to design a learning-based safety value function from the humanoid’s vision and lidar, which in turn can be used to select safe control actions. Importantly, the safety filtering that we design is performed in the low-dimensional space of the three velocity controls of the humanoid, forward velocity, lateral velocity, and the yaw rate, rather than in the high-dimensional joint control space of the robot. This reduces the computational cost and complexity of learning the safety policy.

The remainder of this thesis is organized as follows: Chapter~\ref{chap:preliminaries} introduces the background and preliminary concepts of Hamilton--Jacobi reachability. Chapter~\ref{chap:latent_based_safety_filtering} presents the proposed latent safety filtering framework for humanoid locomotion, including the safety value function, safety policy, training methodology, and simulation results using different sensor modalities and their combinations. Chapter~\ref{chap:sim_to_real} discusses the sim-to-real transfer of the proposed framework and fine-tuning techniques, including its deployment and evaluation on the Unitree G1 humanoid robot. Finally, Chapter~\ref{chap:conclusion and future work} summarizes the main findings and discusses limitations and directions for future work.

\chapter{Preliminaries}
\label{chap:preliminaries}

\section{Hamilton--Jacobi Reachability}

We consider a continuous-time dynamical system of the form:
\begin{equation}
    \dot{x}(t) = f_c(x(t), u(t), d(t)),
    \label{eq:continuous_dynamics}
\end{equation}

where $x(t) \in \mathcal{X} \subseteq \mathbb{R}^n$ is the system state, $u(t) \in \mathcal{U} \subseteq \mathbb{R}^m$ is the control input, and $d(t) \in \mathcal{D} \subseteq \mathbb{R}^p$ is the external disturbance at time $t$. The function $f_c: \mathcal{X} \times \mathcal{U} \times \mathcal{D} \rightarrow \mathbb{R}^n$ describes the continuous-time system dynamics.

\noindent\textbf{Definition 1 (Failure Set \cite{BansalHJReachability}).}
\label{def:failure_set}
Given a continuous safety function
$l: \mathcal{X} \rightarrow \mathbb{R}$,
the failure set for system~\eqref{eq:continuous_dynamics}
is defined as
\begin{equation}
    \mathcal{F} =
    \{x \in \mathcal{X} \mid l(x) \leq 0\},
    \label{eq:failure_set}
\end{equation}
where $l(x)$ represent the safety of the system at state $x$. States satisfying $l(x) > 0$ are considered safe, while states satisfying $l(x) \leq 0$ belong to the failure set.

\noindent\textbf{Definition 2 (Backward Reachable Tube \cite{BansalHJReachability}).}
\label{def:BRT}
Given the failure set $\mathcal{F}$, the backward reachable
tube over the time interval $[0,T]$, where $T>0$, is
defined as
\begin{equation}
    \mathcal{BRT} =
    \left\{
    x \in \mathcal{X} \;\middle|\;
    \forall u(\cdot),\,
    \exists d(\cdot),\,
    \exists \tau \in [0,T]
    \text{ such that }
    \xi(\tau; x,u,d) \in \mathcal{F}
    \right\},
    \label{eq:brt}
\end{equation}
where $\xi(\tau; x,u,d)$ denotes the state reached at time $\tau$ from initial state $x$ at time $0$, with control input $u(\cdot)$ and disturbance $d(\cdot)$.
The backward reachable tube contains all states from which the system cannot avoid the failure set within the given time horizon.

\noindent\textbf{Definition 3 (HJ Safety Value Function \cite{BansalHJReachability}).}
\label{def:HJ_safety_value_function}
Given the continuous-time system in~\eqref{eq:continuous_dynamics}, the HJ safety value function over a time horizon $t>0$ is defined as
\begin{equation}
    V_c(x,t) =
    \inf_{d(\cdot)}
    \sup_{u(\cdot)}
    \min_{\tau \in [0,t]}
    l\bigl(\xi(\tau;x,u,d)\bigr),
    \label{eq:hj_value}
\end{equation}
where $l$ is the safety function defined in
Eq.~\eqref{eq:failure_set} and $\xi(\tau;x,u,d)$
denotes the system trajectory. A positive value of
$V_c(x,t)$ indicates that the system can remain outside
the failure set over the time horizon.

\phantomsection
\label{thm:hj_reachability}
\noindent\textbf{Theorem 1 (HJ Reachability \cite{BansalHJReachability}).}
For the continuous-time system in~\eqref{eq:continuous_dynamics},
the HJ safety value function $V_c(x,t)$ can be obtained
by solving the Hamilton--Jacobi--Isaacs variational inequality
\begin{equation}
    \min\left\{
        \frac{\partial V_c(x,t)}{\partial t}
        + H(x,\nabla_x V_c(x,t)),
        \; l(x)-V_c(x,t)
    \right\}=0,
    \label{eq:hji_vi}
\end{equation}
where the Hamiltonian is defined as
\begin{equation}
    H(x,\nabla_x V_c(x,t)) =
    \max_{u\in\mathcal{U}}
    \min_{d\in\mathcal{D}}
    \nabla_x V_c(x,t)^\top f_c(x,u,d).
    \label{eq:hamiltonian}
\end{equation}
The HJI variational inequality is solved backward in time
from $T$ to $0$, subject to the terminal condition
$V_c(x,T)=l(x)$. The backward reachable tube at time
$t$ can then be represented as $\mathcal{B}(t)=\{x\in\mathcal{X}\mid V_c(x,t)\leq 0\}$.

Given a nominal policy $\pi_{\mathrm{nom}}: \mathcal{X}
\rightarrow \mathcal{U}$, the HJ safety value function
can be used to design a safety filter that modifies the
nominal control input when the system is at risk of entering the failure set. The safety filter selects a control input that maximizes the worst-case rate of change of the safety value function under external disturbances. The optimal safety control policy can be obtained as
\begin{equation}
    \pi_{\mathrm{safe}}(x,t) \in
    \arg\max_{u\in\mathcal{U}}
    \min_{d\in\mathcal{D}}
    \nabla_x V_c(x,t)^\top f_c(x,u,d),
    \label{eq:hj_safe_policy}
\end{equation}
where $\pi_{\mathrm{safe}}(x,t)$ denotes the optimal
safety control policy.

\section{Discrete-Time HJ Reachability}

We consider a discrete-time dynamical system of the form:

\begin{equation}
    s_{t+1} = f_d(s_t,a_t),
    \label{eq:discrete_dynamics}
\end{equation}

where $s_t \in \mathcal{S} \subseteq \mathbb{R}^{n_s}$ is the system state and $a_t \in \mathcal{A} \subseteq \mathbb{R}^{n_a}$ is the control action at time step $t$. The function $f_d: \mathcal{S}\times\mathcal{A}\rightarrow\mathcal{S}$ describes the discrete-time state transition.

\noindent\textbf{Definition 4 (Discrete-Time HJ Safety Value Function \cite{BansalHJReachability,SylviaHJR_survey}).}
Given the deterministic discrete-time system in~\eqref{eq:discrete_dynamics}, let $h: \mathcal{S} \rightarrow \mathbb{R}$ be a safety function that defines the failure set $\mathcal{F} = \{s \in \mathcal{S} \mid h(s) \leq 0\}$. The infinite-horizon HJ safety value function is defined as
\begin{equation}
    V_d(s) =
    \sup_{\pi}
    \inf_{t \geq 0}
    h(s_t^\pi),
    \label{eq:discrete_hj_value}
\end{equation}
where $s_t^\pi$ denotes the state reached at time step $t$ from initial state $s$ under policy $\pi: \mathcal{S} \rightarrow \mathcal{A}$. A positive value of $V_d(s)$ indicates that there exists a policy that can keep the system outside the failure set over the infinite time horizon.

\phantomsection
\label{thm:hj_reachability_2}
\noindent\textbf{Theorem 2 (Discrete-Time HJ Bellman Equation \cite{BansalHJReachability,SylviaHJR_survey}).}
For the discrete-time system in~\eqref{eq:discrete_dynamics}, the infinite-horizon HJ safety value function $V_d(s)$ satisfies the Bellman equation
\begin{equation}
    V_d(s) =
    \min\left\{
        h(s),
        \max_{a\in\mathcal{A}}
        V_d(f_d(s,a))
    \right\},
    \label{eq:discrete_hj_bellman}
\end{equation}
where the first term represents the current safety margin, and the second term represents the maximum achievable safety value at the next time step.

Similar to the continuous-time HJ safety filter
introduced in \hyperref[thm:hj_reachability]{Theorem~1}, the discrete-time HJ safety
value function can be used to construct a safety
control policy. The optimal policy is given by
\begin{equation}
    \pi_{\mathrm{safe}}(s) \in
    \arg\max_{a\in\mathcal{A}}
    V_d(f_d(s,a)),
    \label{eq:discrete_safe_policy}
\end{equation}
where $\pi_{\mathrm{safe}}(s)$ selects the control action
that maximizes the safety value of the next state.

\chapter{Latent-Based Safety Filtering for Humanoid Locomotion using Hamilton-Jacobi Reachability}
\label{chap:latent_based_safety_filtering}

This chapter presents a learning-based safety filtering framework for humanoid locomotion using visual observations. The proposed approach integrates visual representation learning with Hamilton–Jacobi (HJ) reachability to learn a safety value function and a corresponding safety policy for collision avoidance. In addition to vision-based safety filtering, we investigate LiDAR-based safety filtering and LiDAR-Vision fusion to examine how different sensing modalities affect safety performance. We evaluate these approaches in a simulated humanoid locomotion environment, focusing on collision avoidance, navigation efficiency, and safety intervention behavior.

\section{Related Work}
\label{sec:related_work}

HJ reachability has been increasingly applied to safety-critical robot navigation and locomotion. First, efficient reachability-based methods for safe navigation of robots in unknown environments have been introduced in early work by Bajcsy et al. \cite{BajcsyReachability-Based_Framework}. Then, reachability can be used to bound tracking errors between a simplified planning model and a more detailed robot dynamics model in order to enable safe real-time motion planning. Such a framework has been presented by Chen et al. \cite{ChenFaSTrack} for so-called FaSTrack. More recently, safety filtering using HJ reachability has been addressed for a variety of different robotic systems. Borquez et al. \cite{BorquezOnSafety} first presented a unified safety and liveness filtering framework that deals with issues of abrupt control switches while trying to preserve as much as possible the performance of the nominal controller. Later methods also include learning-based and adaptive approaches in order to deal with the increased complexity of the corresponding robots. For example, Lu et al. \cite{RenVB-Com} presented adaptive shielding for dealing with model mismatch in real-world reinforcement learning, and Lin et al. \cite{OneFilter} introduced an observation-conditioned safety filter in order to safely extend the controller generalization to different locomotion controllers on a quadrupedal robot in unknown environments. Finally, hybrid dynamical systems can also be treated using reachability, as has been shown by Borquez et al. \cite{BorquezHJ} for a quadrupedal robot, where safety filtering is done between continuous and discrete control modes. In summary, HJ reachability is becoming more and more applicable for safe robotic motion planning, that is, for safe planning, for dealing with dynamic uncertainties, for controller generalization, and for designing safety filters in general.

Safety filtering for robotic systems has also recently moved from a perception-less approach where the safety of a robot is evaluated solely based on an explicit model of the system to a perception-driven safety filtering approach where the safety of a robot is evaluated directly from high-dimensional sensor data. For the case of vision-based navigation, Toufighi et al. \cite{ToufighiDecisionboundary} proposed to generate safety labels and decision boundaries for safe and unsafe waypoints in a perception-driven fashion using Hamilton–Jacobi reachability. Building upon this work, Nakamura et al. \cite{NakamuraK} integrated a visual representation with a world model and performed HJ reachability in the latent space, whereas Tabbara et al. \cite{TabbaraDesignLatent} systematically studied the impact of the perception encoder and the safety value function of the vision-based safety filtering approach. Seo et al. \cite{seoUncertainty-awareLatentSF} extended the latent space safety filtering approach of Toufighi et al. \cite{ToufighiDecisionboundary} to include the uncertainty of the world model in order to identify potentially unsafe out-of-distribution observations. Although the above approaches relied on Hamilton-Jacobi reachability, other approaches used control barrier functions (CBFs) with perception for the safety of the robotic system. In particular, Abdi et al. \cite{Abdi} proposed to use the vision-based CBFs for the RGB-D observations, whereas Keyumarsi et al. \cite{Keyumarsi} used the LiDAR measurements to synthesize the CBFs via Gaussian process learning. De Sa et al. \cite{DaSa} recently proposed the local barrier functions for safe robot navigation, where the local safety is constructed from the point-cloud geometry. The studies mentioned above investigated safety filtering using different perceptual representations. However, since the control approach, the robot used for the experiments and the corresponding safety performance were all too different between the studies, it was not possible to investigate the influence of the sensing modality in more detail. In contrast, we use a unified learning-based HJ safety filtering framework in our study in order to investigate in detail the influence of the number of sensing modalities on the safety and efficiency of the movement.

In addition, the limitation of single-modality sensing for safety-critical robot navigation is twofold. First, each modality captures unique information about the environment. For example, vision-based sensing can provide rich information regarding the objects, their appearance, and shape; but the geometric measurements cannot be accurately estimated. On the other hand, the LiDAR modality provides direct geometric measurements; but it has limitations such as the angular resolution and blind spots. Thus, for safety-critical tasks, vision and LiDAR can provide complementary information. To this end, Chang et al. \cite{Chang} combined camera, LiDAR, and V2X sensing with risk-aware control barrier functions (CBFs) for dynamic obstacle avoidance. Gravina et al. \cite{Gravina} integrated LiDAR and RGB-D sensing with discrete-time CBFs to enable safe navigation in crowded scenarios. The potential offered by multimodal sensing has been demonstrated in the aforementioned studies; however, the impact of different sensing modalities on learning-based HJ safety filtering for humanoid locomotion is an open issue. Specifically, it is yet to be investigated how the learned representations from the pair of vision-only and LiDAR-only sensing modalities can be combined within a unified safety filtering framework to improve the trade-off between collision avoidance and locomotion efficiency.

\section{Problem Formulation}
\label{sec:problem_formulation}

We consider the collision avoidance problem for humanoid robot navigation. With the deterministic discrete-time system formulation introduced in Chapter~\ref{chap:preliminaries}, the robot is controlled by high-level velocity commands, and a low-level locomotion policy is responsible for generating joint-level actions to track those given velocity commands. Specifically, the high-level control action $a_t \in \mathcal{A}$ is defined as
\begin{equation}
    a_t =
    \begin{bmatrix}
        v_{x,t} & v_{y,t} & \omega_t
    \end{bmatrix}^{\top},
    \label{eq:velocity_action}
\end{equation}
where $v_{x,t}$ and $v_{y,t}$ represent the forward and lateral velocities, and $\omega_t$ denotes the yaw rate. By operating in the high-level velocity command space, the safety filter doesn't need to model the humanoid's full dynamics.

In practical robotic systems, the humanoid obtains its observations from its sensors. We consider three sensing configurations: vision-only, LiDAR-only, and vision--LiDAR fusion. Let $m \in \{\mathrm{V},\mathrm{L},\mathrm{VL}\}$ denote the sensing configuration. The corresponding observation is defined as
\begin{equation}
    o_t^m = g_m(s_t),
    \label{eq:sensor_observation}
\end{equation}
where $g_m$ represents the observation process associated with sensing modality $m$. The observations are mapped into compact latent representations using an encoder $E_m$:
\begin{equation}
    z_t^m = E_m(o_t^m),
    \label{eq:latent_representation}
\end{equation}
where $z_t^m \in \mathcal{Z}_m$ denotes the latent representation.

The safety objective is to determine if the humanoid is safe in the moment. Let $d(s_t)$ denote the minimum distance between the robot and the obstacles in the environment, and let $d_{\mathrm{safe}}$ denote a safety margin. We define the safety function as
\begin{equation}
    h(s_t) = d(s_t) - d_{\mathrm{safe}}.
    \label{eq:humanoid_safety_function}
\end{equation}
So the failure set is 
$
    \mathcal{F} =
    \left\{
        s \in \mathcal{S}
        \mid h(s) \leq 0
    \right\}
    \label{eq:humanoid_failure_set}
$ according to Chapter~\ref{chap:preliminaries}. A state is considered unsafe when the robot enters the failure set. The safety function provides a common safety specification for all sensing configurations, allowing their safety performance to be evaluated under the same criterion. 

To accomplish a navigation task, a nominal controller $\pi_{\mathrm{nom}}$ generates a high-level velocity command $a_t^{\mathrm{nom}}$ to guide the robot toward its goal. But the nominal controller does not explicitly guarantee collision avoidance. Therefore we introduce a learning-based HJ safety filter that evaluates the safety of the nominal action according to sensors and intervenes when necessary. For each sensing configuration $m$, let $Q^m(z_t^m,a_t)$ denote the learned safety value function, and let $\pi_{\mathrm{safe}}^m$ denote the safety filter policy. The applied control action is defined as
\begin{equation}
    a_t =
    \begin{cases}
        a_t^{\mathrm{nom}},
        & Q^m(z_t^m,a_t^{\mathrm{nom}}) \geq 0, \\[4pt]
        \pi_{\mathrm{safe}}^m(z_t^m),
        & Q^m(z_t^m,a_t^{\mathrm{nom}}) < 0.
    \end{cases}
    \label{eq:humanoid_safety_filter}
\end{equation}
The safety filter uses nominal actions when safe and otherwise switches to a safety filter policy based on the HJ reachability objective introduced in Chapter~\ref{chap:preliminaries}.

The goal of this work is to investigate the impact of different sensing modalities on learning-based HJ safety filtering for humanoid locomotion. For this purpose, we design three different modalities for safety filtering, vision-only, LiDAR-only, and vision--LiDAR, in a unified learning setup. The same robot dynamics, controller, safety constraints, and learning setup are used in all of the three different configurations, and thus, the impact of sensing modalities can be specifically investigated. Moreover, it is also investigated whether the combination of visual information and LiDAR information can exploit their complementary information to achieve improved safety while preserving the goal-directed locomotion.

\section{Latent Safety Filtering Framework}
\label{sec:vision_safety_framework}

This section presents the proposed latent safety filtering framework for humanoid locomotion, which integrates visual representation learning, a latent world model, and HJ reachability-based safety learning.

\subsection{Framework Overview}
\label{subsec:framework_overview}

\begin{figure*}[t]
    \centering
    \includegraphics[width=\textwidth]{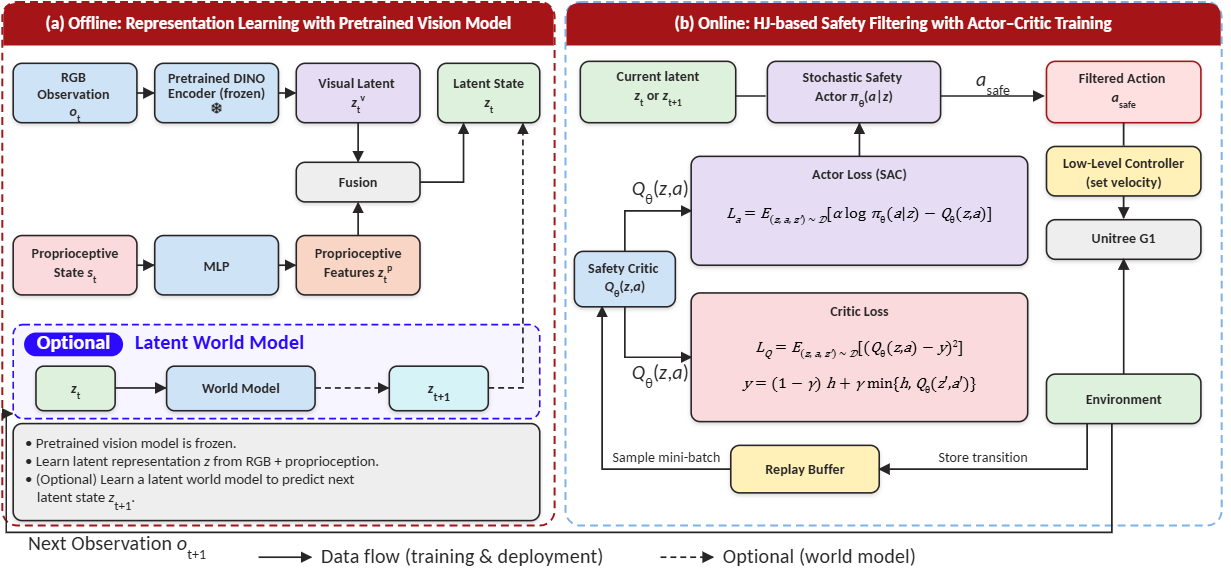}
    \caption{Overview of the proposed latent safety filtering framework, consisting of (a) offline latent representation learning and (b) online HJ-based safety learning with actor-critic training.}
    \label{fig:vision_framework}
\end{figure*}

The overall architecture of the proposed latent safety filtering framework is illustrated in Fig.~\ref{fig:vision_framework}. The framework consists of two stages: offline representation learning and online HJ-based safety learning. In the offline stage, RGB observations and proprioceptive states are encoded and fused into a latent state representation, with an optional world model for predicting future latent states. In the online stage, we use an actor-critic technique to learn the safety policy. Safety critics learn the HJ safety value function from replay buffer transitions, and then a safety actor is optimized using SAC. The safety policy generates high-level velocity commands for collision avoidance, which are then executed by the low-level locomotion controller.

\subsection{Visual Representation and World Model}
\label{subsec:visual_world_model}

At each time step $t$, the RGB observation $o_t^v$ is passed through a frozen DINO network by Caron et al.~\cite{Dino}, $E_{\mathrm{DINO}}$, to obtain the visual features $z_t^v$. The proprioceptive state $p_t$ of the robot is encoded in a set of features $z_t^p$ by a multilayer perceptron (MLP) $E_p$. The visual features and the proprioceptive features are then concatenated to form the latent state representation $z_t = [z_t^v, z_t^p]$. The latent state representation is then used as input to the world model or directly to the HJ-based safety learning framework.

\begin{figure}[!htbp]
    \centering
    \includegraphics[width=\textwidth]{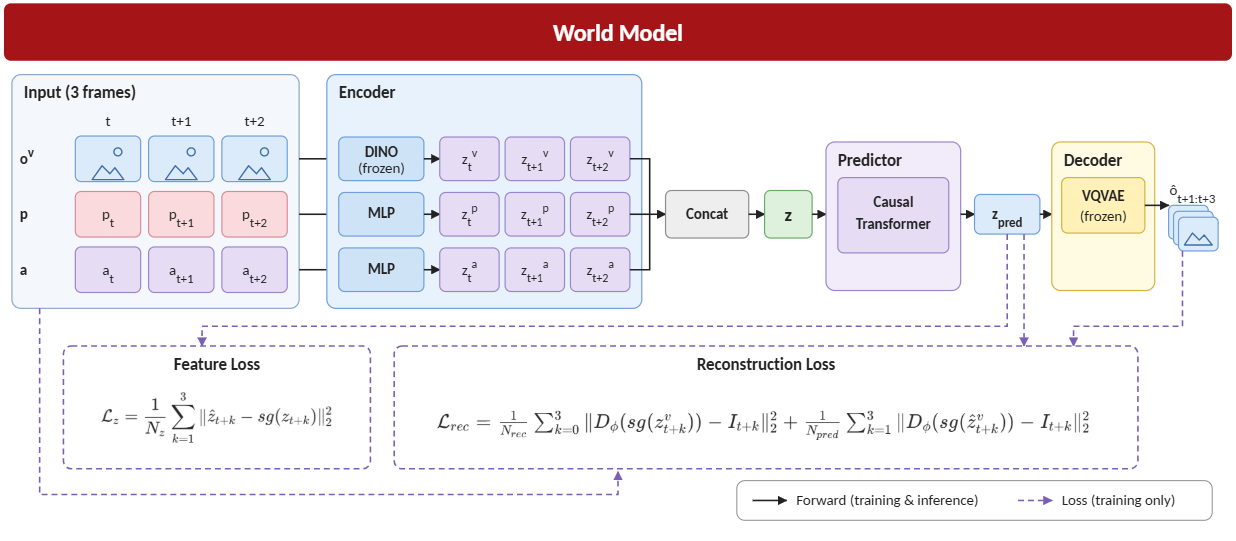}
    \caption{Architecture of the proposed world model, including visual, proprioceptive, and action encoding, latent dynamics prediction using a causal Transformer, and image reconstruction using a VQVAE decoder.}
    \label{fig:world_model}
\end{figure}

To capture the latent dynamics of the robot and its surrounding environment, we train a world model using a six-layer causal Transformer, as illustrated in Fig.~\ref{fig:world_model}. The model takes three consecutive latent states and their corresponding actions as input, and predicts the latent representations of the subsequent three time steps:
\begin{equation}
    (\hat{z}_{t+1}, \hat{z}_{t+2}, \hat{z}_{t+3})
    = W_{\theta}(z_{t:t+2}, a_{t:t+2}),
\end{equation}
where $W_{\theta}$ denotes the world model predictor. The causal attention is restricted to attend only to the current and previous input tokens of the output token. This allows the predictor to learn one-step latent transitions from history.

The world model is trained to minimize the loss of its predictions and the image reconstruction loss. The loss for the predicted features (in terms of mean squared error) against the ground-truth features (visual and proprioceptive) is:
\begin{equation}
    \mathcal{L}_{z}
    = \frac{1}{N_z}
    \sum_{k=1}^{3}
    \left\|
    \hat{z}_{t+k}
    - \operatorname{sg}(z_{t+k})
    \right\|_2^2,
\end{equation}
where $N_z$ is a normalization constant and $\operatorname{sg}(\cdot)$ denotes the stop-gradient operation.

To preserve visual info, the VQ-VAE decoder $D_{\phi}$ in the feature extractor is used to reconstruct RGB observations from the encoded and predicted visual features in order to preserve visual information.

\begin{equation}
\mathcal{L}_{\mathrm{rec}}
=
\frac{1}{N_{\mathrm{rec}}}
\sum_{k=0}^{3}
\left\|
D_{\phi}(\operatorname{sg}(z_{t+k}^{v})) - I_{t+k}
\right\|_2^2
+
\frac{1}{N_{\mathrm{pred}}}
\sum_{k=1}^{3}
\left\|
D_{\phi}(\operatorname{sg}(\hat{z}_{t+k}^{v})) - I_{t+k}
\right\|_2^2.
\end{equation} where $z_{t+k}^{v}$ and $\hat{z}_{t+k}^{v}$ are the encoded visual features for time step $t+k$ and the predicted visual features for time step $t+k$, respectively. $I_{t+k}$ is the ground-truth RGB observation for time step $t+k$. $N_{\mathrm{rec}}$ and $N_{\mathrm{pred}}$ are constants that normalize the loss by the number of elements in each part of the loss. The stop-gradient operation prevents the loss from updating the visual encoder and world model predictor.

The overall training objective is
\begin{equation}
    \mathcal{L}_{\mathrm{total}}
    = \mathcal{L}_{z} + \mathcal{L}_{\mathrm{rec}}.
\end{equation}

During training, the encoders remain frozen, while the causal Transformer predictor is optimized through the latent prediction loss and the VQ-VAE decoder is optimized through the reconstruction loss. The trained world model can then predict the one-step-ahead latent state from historical latent representations and actions, providing a predicted future state as input to the safety filter.

\subsection{Learning the HJ Safety Value Function}
\label{subsec:hj_safety_learning}

We learn a safety action-value function $Q_\theta(z_t,a_t)$ in the latent space to evaluate the long-term safety of executing action $a_t$ under the latent state $z_t$. Following the HJ reachability formulation introduced in Chapter~\ref{chap:preliminaries}, we use the safety function $h(o_t)$ defined in Eq.~\eqref{eq:humanoid_safety_function} as the supervision signal for learning the HJ safety action-value function. Next, to learn the HJ safety value function, we adopt an off-policy actor-critic approach with the safety critics, $Q_{\theta}(z_t,a_t)$. It's trained using transitions $(z_t,a_t,h(o_t),z_{t+1})$ sampled from a replay buffer. The critics are optimized by minimizing the mean squared Bellman error:
\begin{equation}
    \mathcal{L}_Q =
    \mathbb{E}_{(z,a,h,z')\sim\mathcal{D}}
    \left[
        \left(Q_{\theta}(z,a)-y\right)^2
    \right],
    \label{eq:hj_critic_loss}
\end{equation}
where $\mathcal{D}$ denotes the replay buffer and $y$ is the target safety value computed using the discounted HJ Bellman formulation:
\begin{equation}
    y_t = (1-\gamma)h(z_t)
    + \gamma \min\left\{
        h(z_t),
        Q_{\bar{\theta}}(z_{t+1},a_{t+1})
    \right\},
    \label{eq:hj_bellman_target}
\end{equation}
where $\gamma \in [0,1)$ is the discount factor, $\bar{\theta}_i$ denotes the target critic parameters, and $a_{t+1}$ is sampled from an actor.

At the same time, the stochastic safety actor $\pi_\phi(a \mid z)$ is also optimized by using the Soft Actor-Critic (SAC) objective to maximize the learned safety value, using entropy regularization to encourage exploration. We optimize the actor in the following way:
\begin{equation}
\mathcal{L}_{\pi} =
\mathbb{E}_{z \sim \mathcal{D},\, a \sim \pi_\phi(\cdot \mid z)}
\left[
\alpha \log \pi_\phi(a \mid z)
- Q_{\theta}(z,a)
\right],
\label{eq:hj_actor_loss}
\end{equation}
Here, the coefficient $\alpha$ balances the trade-off between the entropy term and the safety term. By using this actor-critic learning, the safety actor $\pi_{\phi}(a|z)$ is optimized to find a set of high-level commands that can maximize the estimated long-term safety for the robot in a given state $z$.

\subsection{Safety Filter and Policy Switching}
\label{subsec:safety_policy_switching}

The safety filter evaluates the nominal velocity command using the learned HJ safety value function, $Q(z_t,a_t) = Q_{\theta}(z_t,a_t)$. The filtered action is then determined by
\begin{equation}
    a_t^{\mathrm{SF}} =
    \begin{cases}
        \pi_\phi(z_t), & Q(z_t,a_t^{\mathrm{nom}}) < \tau,\\
        a_t^{\mathrm{nom}}, & \text{otherwise},
    \end{cases}
    \label{eq:safety_filter_switching}
\end{equation}
where $a_t^{\mathrm{nom}} = \pi_{\mathrm{nom}}(z_t)$ is the nominal velocity command and $\tau$ is the safety threshold. The complete switching procedure is summarized in Algorithm~\ref{alg:safety_filter}:

\begin{algorithm}[!htbp]
\caption{HJ-Based Safety Filtering and Policy Switching}
\label{alg:safety_filter}

\KwIn{Latent state $z_t$, nominal controller $\pi_{\mathrm{nom}}$,
safety actor $\pi_\phi$, safety critic $Q$,
world model $W_\eta$, threshold $\tau$, evaluation mode $m$}
\KwOut{Filtered velocity command $a_t$}

$a_t^{\mathrm{nom}} \gets \pi_{\mathrm{nom}}(z_t)$\;

\eIf{$m = \mathrm{lookahead}$}{
    $\hat{z}_{t+1} \gets W_\eta(z_t,a_t^{\mathrm{nom}})$\;
    $q_t \gets \max_{a \in \mathcal{A}} Q(\hat{z}_{t+1},a)$\;
}{
    $q_t \gets Q(z_t,a_t^{\mathrm{nom}})$\;
}

\eIf{$q_t < \tau$}{
    $a_t \gets \pi_\phi(z_t)$\;
}{
    $a_t \gets a_t^{\mathrm{nom}}$\;
}
\Return{$a_t$}\;
\end{algorithm}

\section{LiDAR-Based and Multimodal Safety Filtering}
\label{sec:lidar_multimodal_filtering}

LiDAR-based safety filtering is considered to evaluate how geometric distance measurements can support safe locomotion, information which vision-based sensing is lacking. We further investigate combining LiDAR and vision to examine whether using both sensing modalities improves safety filtering performance by taking advantage of their complementary properties.

For LiDAR-based safety filtering, we adopt Sparse Point-Voxel U-Net (SpUNet) by Choy et al.~\cite{SpUNet} as the LiDAR encoder to extract features from the input point cloud. The resulting latent representation, $z_{\mathrm{LiDAR}}$, is used as the observation input to the safety value network and safety policy network described in subsection~\ref{subsec:framework_overview} and subsection~\ref{subsec:visual_world_model}. The remaining components of the safety filtering framework, including the training procedure and switching mechanism, remain the same from the vision-based approach.

For Vision--LiDAR fusion, we employ the DINO vision encoder and the SpUNet LiDAR encoder to extract features from the sensor data of both modalities. The two feature representations are then concatenated to form the joint feature representation of the Vision--LiDAR fusion:
\begin{equation}
z_{\mathrm{fusion}} =
\left[z_{\mathrm{vision}}|| z_{\mathrm{LiDAR}}\right].
\end{equation} This fused representation is used instead of the single-modality latent space within the safety filtering framework. The rest of the network remains unchanged.

\section{Simulation Experiments}
\label{sec:simulation_experiments}

In this section, we evaluate the proposed safety filtering framework in simulation to examine its ability to prevent collisions while maintaining navigation performance. We also compare different sensing modalities to investigate how visual and geometric information affect safety filtering performance and robot navigation behavior.

\subsection{Experimental Setup}
\label{subsec:experimental_setup}

\noindent\textbf{1. Simulation Environment and Robot Platform}

We conduct simulation experiments using NVIDIA Isaac Sim and Isaac Lab with the PhysX physics engine. The simulation is built upon the Isaac Lab velocity-tracking task for the Unitree G1 humanoid robot, with a physics time step of 0.005 s. We use the full G1 robot model and a pretrained PPO locomotion policy provided by Isaac Lab through RSL-RL. The locomotion policy takes three-dimensional velocity commands $(v_x, v_y, \omega)$ and generates joint position targets to control the robot. This lower-level PPO policy remains frozen throughout our experiments.

In our simulation, we used a reconstructed laboratory scene. For visual rendering of this scene, we employed 3D Gaussian Splatting (3DGS) \cite{kerbl2023gaussian}. For the reconstruction of the environment, we recorded videos of the real laboratory from different views. From these videos, we extracted the frames of the images and processed them using COLMAP \cite{schonberger2016structure} to compute the poses of the cameras and to reconstruct the scene. We then used this reconstructed scene to train a 3DGS model in order to reproduce the views of the laboratory. We converted the reconstructed scene to USDZ format using NVIDIA 3DGRUT \cite{wu20253dgut} in order to integrate it into the environment of Isaac Sim. In addition, we used 2D Gaussian Splatting (2DGS) \cite{huang2024gaussian} in order to reconstruct a surface mesh from the input of the TSDF-fusion. This surface mesh can then be used as collision geometry for the static environment.

In order to test the obstacle avoidance feature, we add dynamic physical obstacles with collision geometry to the environment that has been reconstructed for the experiment. To generate the obstacle’s mesh from the reference images, we employ the Meta 3D AssetGen tool \cite{siddiqui2024assetgen,sam3d2025} and export the generated mesh in GLB format. The USD asset is then converted from the GLB file using Isaac Lab’s MeshConverter, where we perform the convex decomposition to generate the collision geometry for the obstacle, and we also set the rigid-body properties for the obstacle. Each of the obstacles is defined as a dynamic rigid body in the PhysX simulation and the robot can interact physically with the obstacle. For the experiments that have been conducted so far, we use blue bins with a mass of 2 kg and a scale factor of 0.65. There are up to two bins, which are placed near the nominal path of the robot and whose positions and numbers are varied for each episode. The positions of the bins are changed for each new episode, whereas the rigid-body and collision properties of the bins remain unchanged.

\begin{figure}[htbp]
    \centering
    \includegraphics[width=0.85\textwidth]{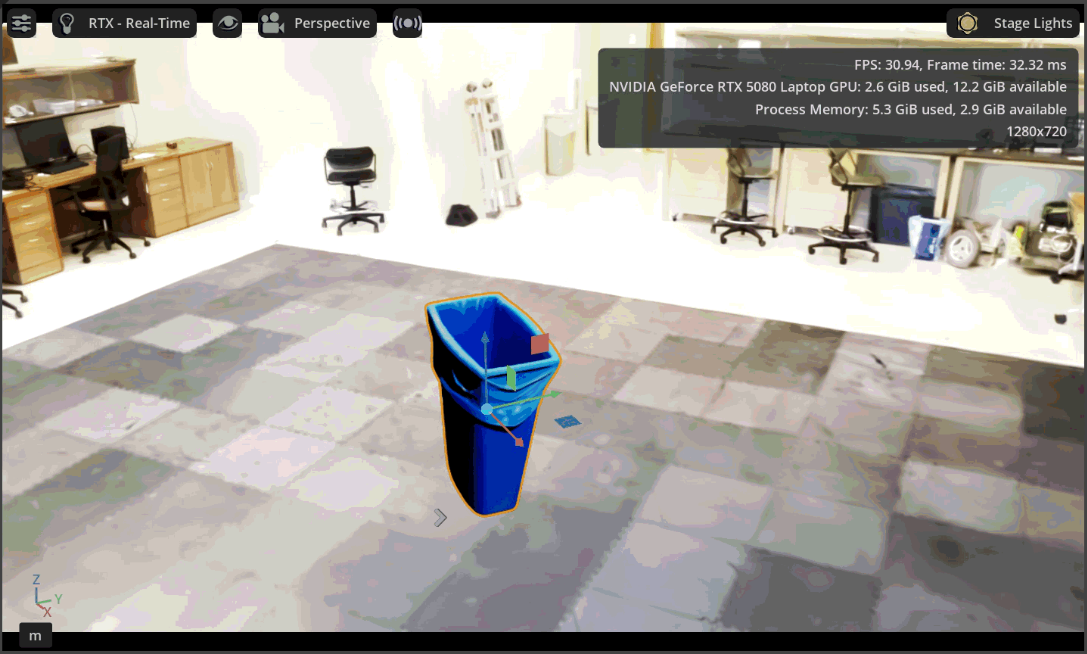}
    \caption{Simulation environment reconstructed using 3D Gaussian Splatting, with a blue bin introduced as a physical obstacle for safety filtering experiments.}
    \label{fig:simulation_scene}
\end{figure}

\noindent\textbf{2. Sensor Configuration}

The simulated Unitree G1 is equipped with an RGB camera and LiDAR sensors mounted on its head. The RGB camera is modeled as an RTX pinhole camera with a focal length of 24 mm and a horizontal aperture of 20.955 mm. RGB observations are collected at 15 Hz and resized to $192 \times 256$ pixels for data storage and processing.

Two LiDAR configurations are used for different purposes. The first is a 45-channel RayCaster LiDAR with a vertical field of view of $[-90^\circ, 90^\circ]$, a horizontal field of view of $[-90^\circ, 90^\circ]$, and an angular resolution of $2^\circ$. This configuration is used to compute the safety supervision signal, which is the closest distance from the robot to its nearest obstacle. The second configuration simulates a Mid-360 LiDAR with 85 vertical channels, a vertical field of view of $[-52^\circ, 7^\circ]$, a $360^\circ$ horizontal field of view, and a maximum sensing range of 40 m, which mimics the LiDAR setting for the real G1 robot and it serves as a LiDAR observation to the LiDAR encoder for LiDAR-based safety filtering. Both configurations operate at 7 Hz, with ray casting performed against the physical obstacle meshes rather than the visual laboratory background.

\noindent\textbf{3. Nominal Waypoint Controller}

We implement a waypoint-based nominal controller in order to generate velocity commands for the humanoid to walk towards a certain target position, described in Algorithm~\ref{alg:waypoint_controller}. For each step of the control loop the displacement between the current waypoint and the target waypoint is transformed into the local frame of the humanoid in order to compute the desired planar velocity and the corresponding yaw rate. If the target waypoint is outside the forward-facing region of the humanoid, the linear velocity is decreased in order to first rotate the robot towards the target waypoint. In order to avoid temporal discontinuities of the generated velocity commands, we also apply temporal smoothing. The resulting nominal action $\mathbf{a}_{\mathrm{nom}} = (v_x, v_y, \omega)$ is limited to $\pm 0.5$ m/s for the linear velocities $v_x, v_y$ and $\pm 1$ rad/s for the yaw rate $\omega$. The lower-level PPO locomotion policy is then these velocity command to compute the joint positions for the nominal action $\mathbf{a}_{\mathrm{nom}}$. A waypoint is considered to have been reached by the humanoid if the target position is within a distance of 0.1 m.

\begin{algorithm}[htbp]
\caption{Waypoint-based Locomotion higher-level controller}
\label{alg:waypoint_controller}
\KwIn{Current robot state $(x_t, y_t, \psi_t)$, target waypoint $w = (x_w, y_w)$}
\KwOut{Velocity command $(v_x, v_y, \dot{\psi})$}
\KwIn{Maximum speed $v_{\max}$, yaw gain $k_{\text{yaw}}$, yaw rate limit $\dot{\psi}_{\max}$,
facing threshold $\theta_{\text{th}}$, dead-zone width $\theta_{\text{dz}}$}

Compute waypoint displacement in world frame:\;
$\Delta x \leftarrow x_w - x_t$, \quad $\Delta y \leftarrow y_w - y_t$\;

Compute distance to waypoint:\;
$d \leftarrow \sqrt{\Delta x^2 + \Delta y^2}$\;

\If{$d < d_{\text{stop}}$}{
    Stop or switch to next waypoint\;
}

Transform displacement to robot local frame:\;
$\Delta x_{\text{local}} \leftarrow \cos(\psi_t)\Delta x + \sin(\psi_t)\Delta y$\;
$\Delta y_{\text{local}} \leftarrow -\sin(\psi_t)\Delta x + \cos(\psi_t)\Delta y$\;

Normalize local direction vector:\;
$\mathbf{u} \leftarrow \frac{1}{\|\left(\Delta x_{\text{local}}, \Delta y_{\text{local}}\right)\|}
(\Delta x_{\text{local}}, \Delta y_{\text{local}})$\;

Compute desired planar velocity:\;
$v_x^* \leftarrow v_{\max} \cdot u_x$, \quad
$v_y^* \leftarrow v_{\max} \cdot u_y$\;

Compute velocity-facing angle:\;
$\theta_v \leftarrow \mathrm{atan2}(v_y^*, v_x^*)$\;

Normalize $\theta_v$ to $[-\pi, \pi]$ using shortest-angle mapping\;

Convert to degrees:\;
$\theta_{\text{deg}} \leftarrow |\theta_v| \cdot 180 / \pi$\;

\eIf{$\theta_{\text{deg}} \leq \theta_{\text{th}}$}{
    \tcp{Target is within forward-facing region}
    $v_x \leftarrow v_x^*$, \quad $v_y \leftarrow v_y^*$\;
    $\dot{\psi}^* \leftarrow k_{\text{yaw}} \cdot \theta_v$\;
}{
    \tcp{Target is outside forward-facing region}
    Reduce linear velocity:\;
    $v_x \leftarrow v_{\text{slow}}$, \quad $v_y \leftarrow 0$\;

    \eIf{$\theta_{\text{deg}} \geq 180 - \theta_{\text{dz}}$}{
        \tcp{Back dead-zone: force fixed rotation}
        $\dot{\psi}^* \leftarrow \dot{\psi}_{\max}$\;
    }{
        \tcp{Regular misalignment: proportional yaw control}
        $\dot{\psi}^* \leftarrow \mathrm{clip}(k_{\text{yaw}}\cdot\theta_v,
        -\dot{\psi}_{\max}, \dot{\psi}_{\max})$\;
    }
}

Apply temporal smoothing to yaw rate:\;
$\dot{\psi}_t \leftarrow (1-\alpha)\dot{\psi}_{t-1} + \alpha\dot{\psi}^*$\;

Apply temporal smoothing to linear velocity:\;
$v_x \leftarrow (1-\beta)v_{x,t-1} + \beta v_x$\;
$v_y \leftarrow (1-\beta)v_{y,t-1} + \beta v_y$\;

\Return $(v_x, v_y, \dot{\psi}_t)$\;
\end{algorithm}

\noindent\textbf{4. Scenarios and Baselines}

The safety filtering framework is evaluated in a set of randomized navigation scenarios, in a rectangular environment of size $7 \times 6$ m. The start and goal positions are sampled uniformly at random within the environment boundaries. The number of blue bins in a scenario is zero, one or two, with probabilities $10\%$, $45\%$ and $45\%$ respectively. The obstacles are placed near the start-to-goal path at locations.

We evaluate a simple waypoint controller without safety filtering in comparison to safety filtering based on vision and LiDAR in a large number of randomized scenarios within a fixed rectangular test area of 7 m × 6 m. We use the same locomotion policy and the same simple nominal controller in all cases. The vision-based safety filtering is based on feature extraction from RGB images, while the LiDAR-based safety filtering is based on encoded point cloud features. All the safety filter methods will have the same set of scenarios configuration for evaluation.

\subsection{Evaluation Metrics}
\label{subsec:evaluation_metrics}

We evaluate the safety filtering performance using five metrics. Each metric is computed for individual trials and then averaged across trials unless otherwise specified.

\begin{enumerate}
\item \textbf{Success Rate:} The ratio of the number of trials in which the robot completes all the waypoints of a trial without any collisions.

\item \textbf{Collision Rate:} The rate of trajectories with at least one collision for the robot. The number of such collisions for a given trajectory is counted at every observation step. A collision for a given link of the robot (except for the ankle roll link) is detected if the absolute value of any contact force component exceeds 0.1 N at the given observation step. The average collision rate is then calculated as the number of all trajectories with collision divided by the number of all trajectories that were evaluated.

\item \textbf{Average Velocity:} Average horizontal velocity of the robot in all test cases. The average speed is computed as the planar distance of the robot’s center divided by the time between two observations (sampling rate 15 Hz). The average is first computed for each test case and then for all test cases.

\item \textbf{Filter Activation Rate:} The average fraction of control steps in which the safety filter’s action for the current time step is different from the nominal action. The switching mode’s safety value $Q(z, a_{\mathrm{nom}})$ needs to fall below the switching threshold for the intervention to occur.

\item \textbf{Minimum Obstacle Distance:} Average of the minimum horizontal distance of the base of the robot from the center of the nearest obstacle in all considered trials (both successful and failed ones). For the obstacle-free trials this value is not considered.
\end{enumerate}

\subsection{Safety Filtering and World Model Performance}
\label{subsec:safety_performance}

In this section, we evaluate the performance of the safety filtering approaches in simulation using the five evaluation metrics. We first compare the quantitative results across different sensing modalities, followed by an analysis of their navigation behaviors and the performance of the visual world model.

\noindent\textbf{Safety Filtering Performance}

Table~\ref{tab:safety_filter_comparison} compares the nominal waypoint controller with three safety filtering approaches without a world model in simulation: LiDAR-only SF, vision-only SF, and LiDAR--Vision SF. The nominal controller achieved a success rate of 0.55 and a collision rate of 0.49 without safety filtering. All three safety filters improved the success rate and reduced the collision rate. Both the LiDAR-only SF and LiDAR--Vision SF achieved the highest success rate of 0.79, while the vision-only SF achieved a lower success rate of 0.72. The LiDAR--Vision SF achieved the lowest collision rate of 0.18, followed by the LiDAR-only SF at 0.20 and the vision-only SF at 0.30. These results indicate that combining LiDAR and vision further improves collision avoidance while maintaining the same success rate as the LiDAR-only SF. In terms of average velocity, the LiDAR--Vision SF achieved the higher value of 0.418 m/s, exceeding the nominal controller (0.402 m/s), while the LiDAR-only SF and vision-only SF achieved slightly lower average velocities of 0.391 m/s and 0.393 m/s, respectively. Regarding the filter activation rate, the LiDAR-only SF had the highest value of 0.373, followed by the LiDAR--Vision SF at 0.289 and the vision-only SF at 0.232. Furthermore, the LiDAR--Vision SF maintained the largest minimum obstacle distance of 0.837 m, followed by the LiDAR-only SF at 0.782 m and the vision-only SF at 0.693 m. In summary, the LiDAR--Vision SF achieved better collision avoidance while maintaining a higher average velocity, requiring fewer interventions than the LiDAR-only SF, and maintaining a greater distance from obstacles than either single-modality safety filter.

\begin{table}[!htbp]
\centering
\caption{Comparison of safety filtering performance in simulation}
\label{tab:safety_filter_comparison}
\resizebox{\textwidth}{!}{%
\begin{tabular}{lccccc}
\hline
\textbf{Filter} &
\shortstack{\textbf{Success}\\\textbf{Rate} $\uparrow$} &
\shortstack{\textbf{Collision}\\\textbf{Rate} $\downarrow$} &
\shortstack{\textbf{Average}\\\textbf{Velocity (m/s)} $\uparrow$} &
\shortstack{\textbf{Filter Activation}\\\textbf{Rate} $\downarrow$} &
\shortstack{\textbf{Minimum Obstacle}\\\textbf{Distance (m)} $\uparrow$} \\
\hline
No Filter & 0.55 & 0.49 & $0.402 \pm 0.067$ & $\mathbf{0.000 \pm 0.000}$ & $0.544 \pm 0.364$ \\
Lidar-only SF & 0.79 & 0.20 & $0.391 \pm 0.089$ & $0.373 \pm 0.228$ & $0.782 \pm 0.325$ \\
Vision-only SF & 0.72 & 0.30 & $0.393 \pm 0.075$ & $0.232 \pm 0.208$ & $0.693 \pm 0.312$ \\
Lidar-Vision SF & 0.79 & \textbf{0.18} & $0.418 \pm 0.060$ & $0.289 \pm 0.182$ & $\mathbf{0.837 \pm 0.297}$ \\
Lidar-only WM SF & 0.75 & 0.23 & $0.345 \pm 0.101$ & $0.446 \pm 0.263$ & $0.710 \pm 0.307$ \\
Vision-only WM SF & 0.74 & 0.27 & $\mathbf{0.432 \pm 0.060}$ & $0.185 \pm 0.147$ & $0.695 \pm 0.345$ \\
Lidar-Vision WM SF & \textbf{0.82} & 0.19 & $0.416 \pm 0.064$ & $0.322 \pm 0.225$ & $0.761 \pm 0.336$ \\
\hline
\end{tabular}%
}
\end{table}

We also compared the performance of the three safety filtering methods with world models. For the vision-only SF, the addition of a world model increased the success rate from 0.72 to 0.74 and reduced the collision rate from 0.30 to 0.27. It also achieved the highest average velocity of $0.432$ m/s and the lowest filter activation rate of $0.185$ among all six safety filtering methods. In contrast, the LiDAR-only WM SF decreased the success rate from 0.79 to 0.75 and increased the collision rate from 0.20 to 0.23, while the addition of a world model to the LiDAR--Vision SF increased the success rate from 0.79 to 0.82 and only slightly increased the collision rate from 0.18 to 0.19. Hence, the benefits of world models might differ depending on the sensing modalities. While world models provide some improvements for vision-based safety filtering, their benefits for methods utilizing LiDAR information are less significant.

\begin{figure}[H]
    \centering
    \begin{minipage}[t]{0.48\textwidth}
        \centering
        \includegraphics[width=\linewidth]{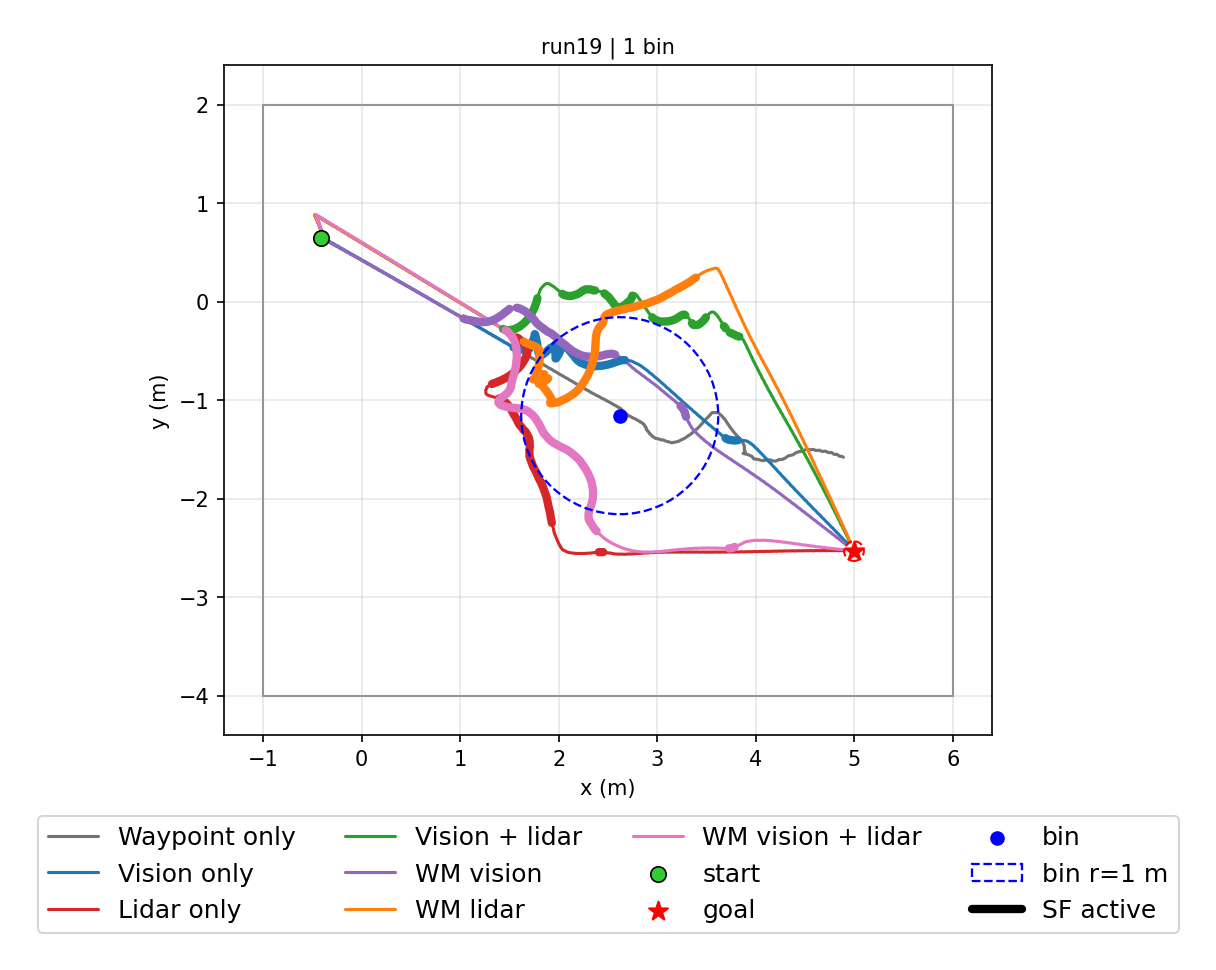}
        \par (a) One obstacle
    \end{minipage}
    \hfill
    \begin{minipage}[t]{0.48\textwidth}
        \centering
        \includegraphics[width=\linewidth]{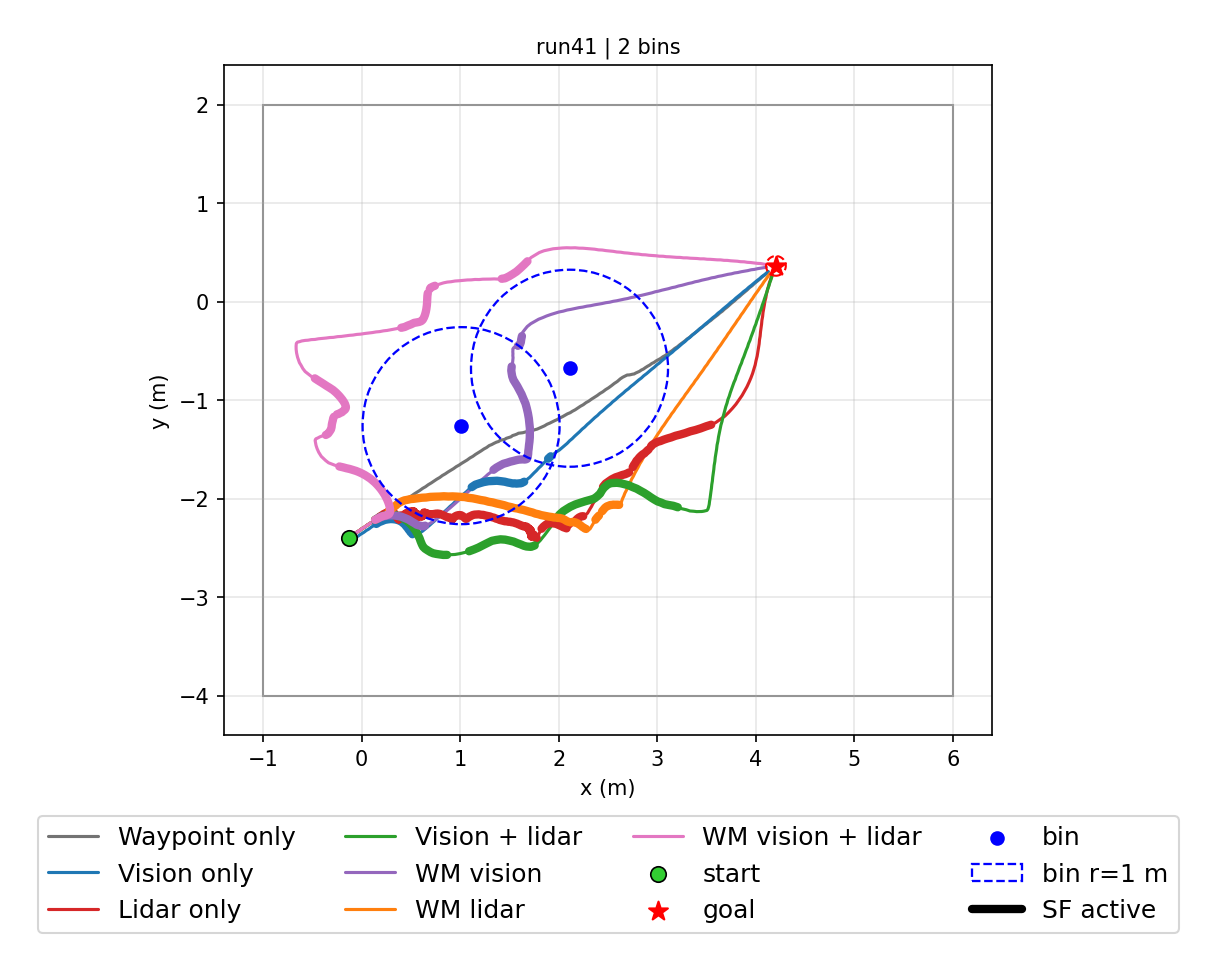}
        \par (b) Two obstacles
    \end{minipage}

    \caption{Comparison of robot navigation trajectories under two obstacle configurations: (a) one obstacle and (b) two obstacles. Trajectories are shown for the nominal waypoint controller, vision-only safety filter, LiDAR-only safety filter, and vision--LiDAR fusion safety filter. Dashed circles indicate the 1 m safety boundaries around obstacles, while thicker trajectory segments indicate safety filter intervention.}
    \label{fig:simulation_trajectories}
\end{figure}

\begin{figure}[H]
    \centering
    \includegraphics[width=0.4\textwidth]{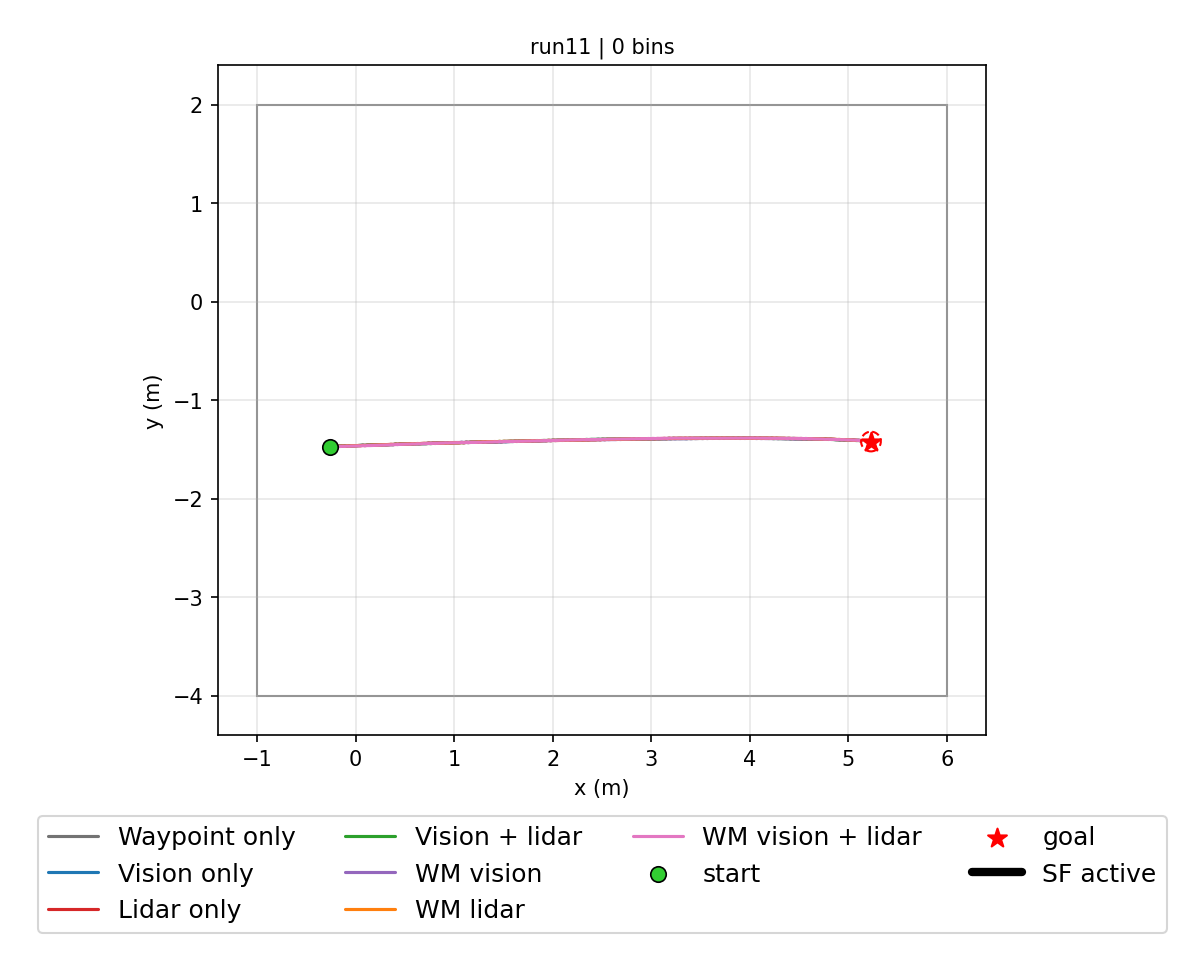}
    \caption{Comparison of robot navigation trajectories in an obstacle-free environment. All safety filtering methods produce trajectories similar to that of the nominal waypoint controller.}
    \label{fig:simulation_no_obstacles}
\end{figure}

The Figure~\ref{fig:simulation_trajectories} shows the trajectories of the nominal waypoint controller as well as the three safety filtering approaches for different configurations of obstacles in the environment. The trajectories are similar in the obstacle-free environment (see Fig.~\ref{fig:simulation_no_obstacles}). This scenario was a control group to demonstrate that the safety filtering approaches do not only remember the sensors, but actually learn to detect obstacles for safe navigation.

For one obstacle (Fig.~\ref{fig:simulation_trajectories}a) the nominal path to the goal is modified in order to avoid it by the safety filters. The vision-only safety filter changes the path just a little bit from the nominal line and nearly collide with the obstacle. The paths modified by the LiDAR-only and LiDAR–Vision safety filters deviate more from the nominal path to avoid the single obstacle, although the trajectory under the LiDAR-only safety filter enters the failure zone. In the two-obstacle scenario (Fig.~\ref{fig:simulation_trajectories}b), the safety filtering methods exhibit more distinct navigation behaviors. The vision-only safety filter produces a trajectory close to the nominal path and only steering when it got really close to the obstacle. In contrast, the LiDAR-only and LiDAR--Vision safety filters guide the robot around the right side of the obstacles. Among the world-model-based methods, the vision-only safety filter entered to the failure zone and trying to avoid collision by pass through in between of the two obstacles. The LiDAR--Vision WM safety filter takes a different route around the left side of the obstacles. These trajectories illustrate how different sensing modalities and the incorporation of world models influence the obstacle avoidance behavior of the robot.

\noindent\textbf{World Model Performance}

\begin{figure}[H]
    \centering
    \begin{minipage}[t]{0.32\textwidth}
        \centering
        \includegraphics[width=\linewidth]{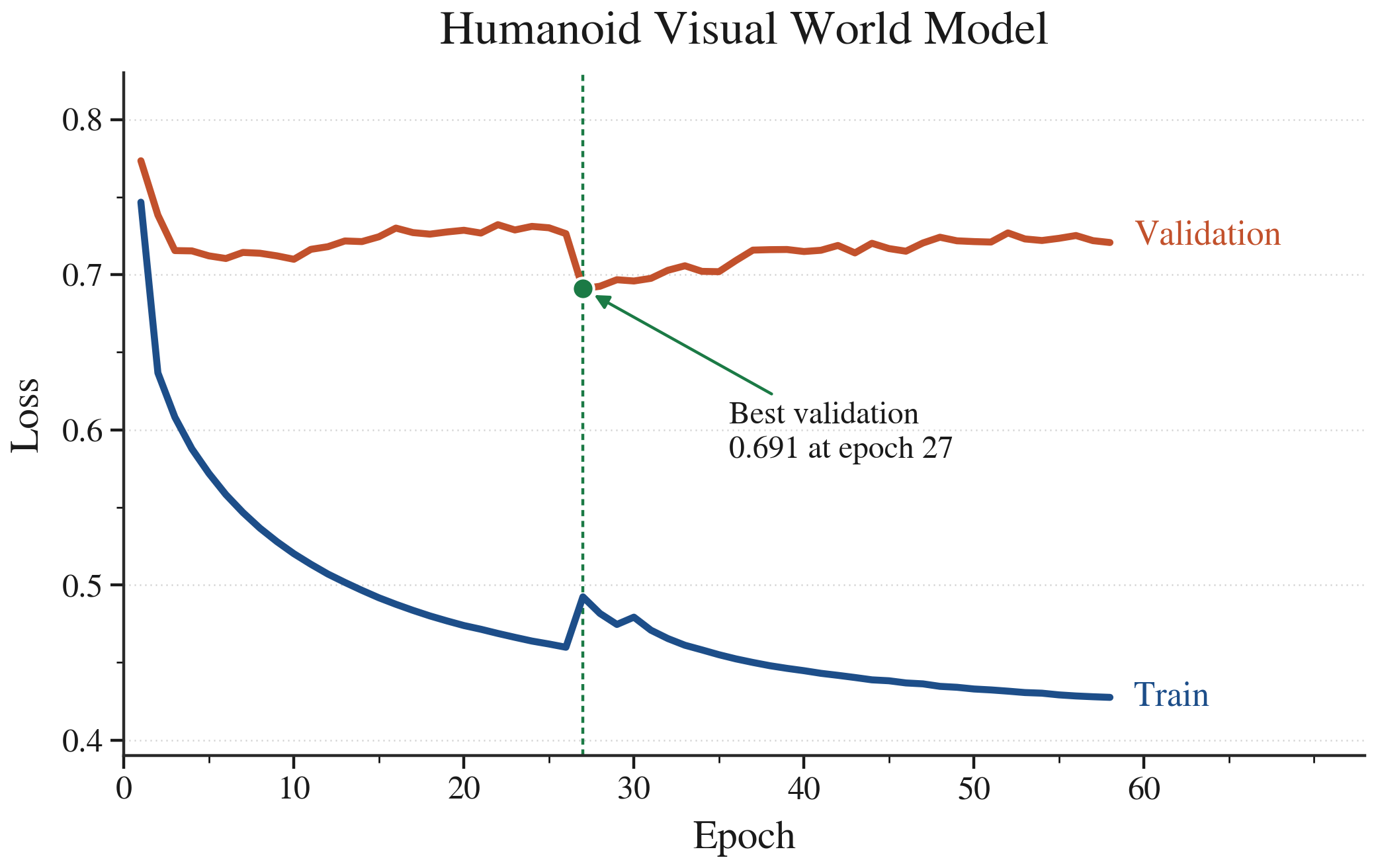}
        \par (a) Vision-based world model
    \end{minipage}
    \hfill
    \begin{minipage}[t]{0.32\textwidth}
        \centering
        \includegraphics[width=\linewidth]{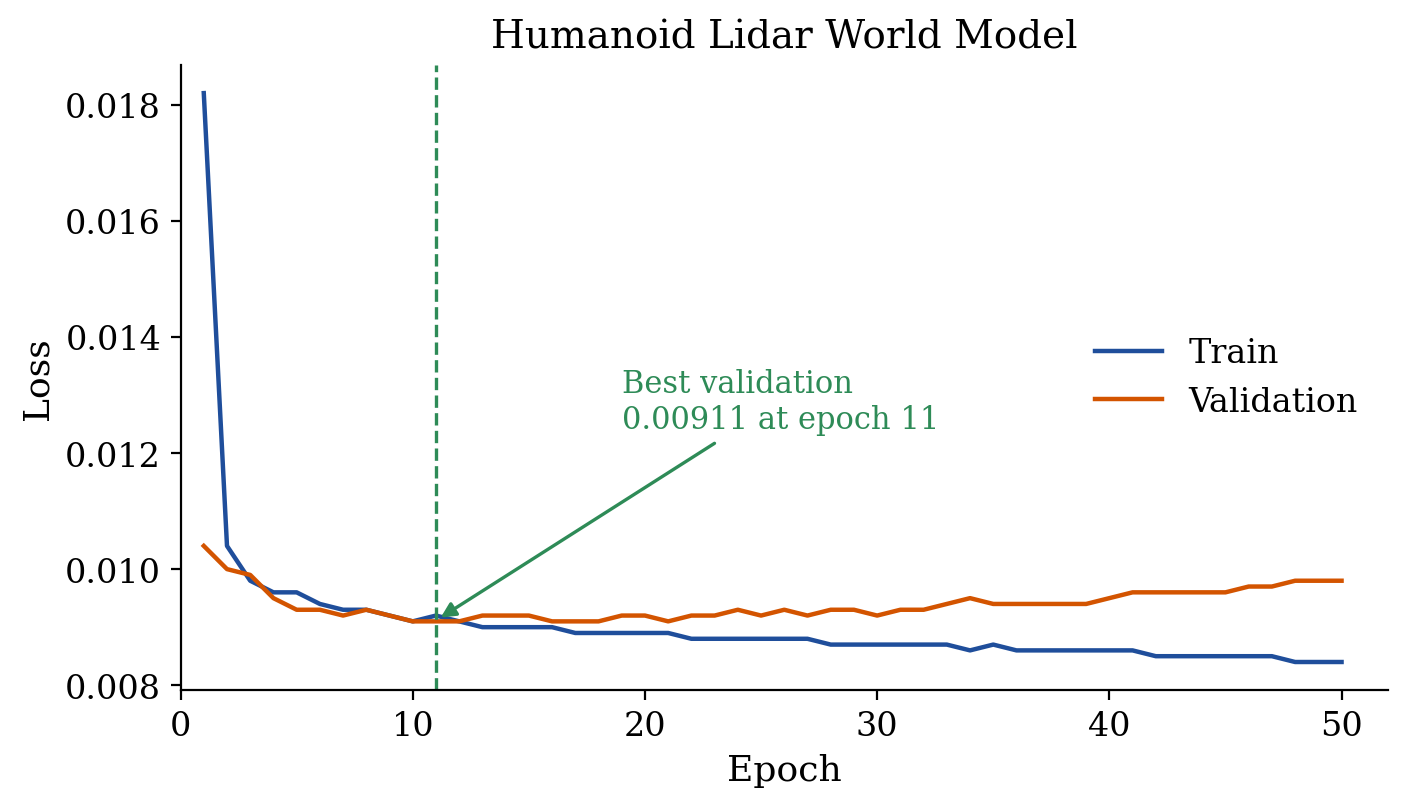}
        \par (b) LiDAR-based world model
    \end{minipage}
    \hfill
    \begin{minipage}[t]{0.32\textwidth}
        \centering
        \includegraphics[width=\linewidth]{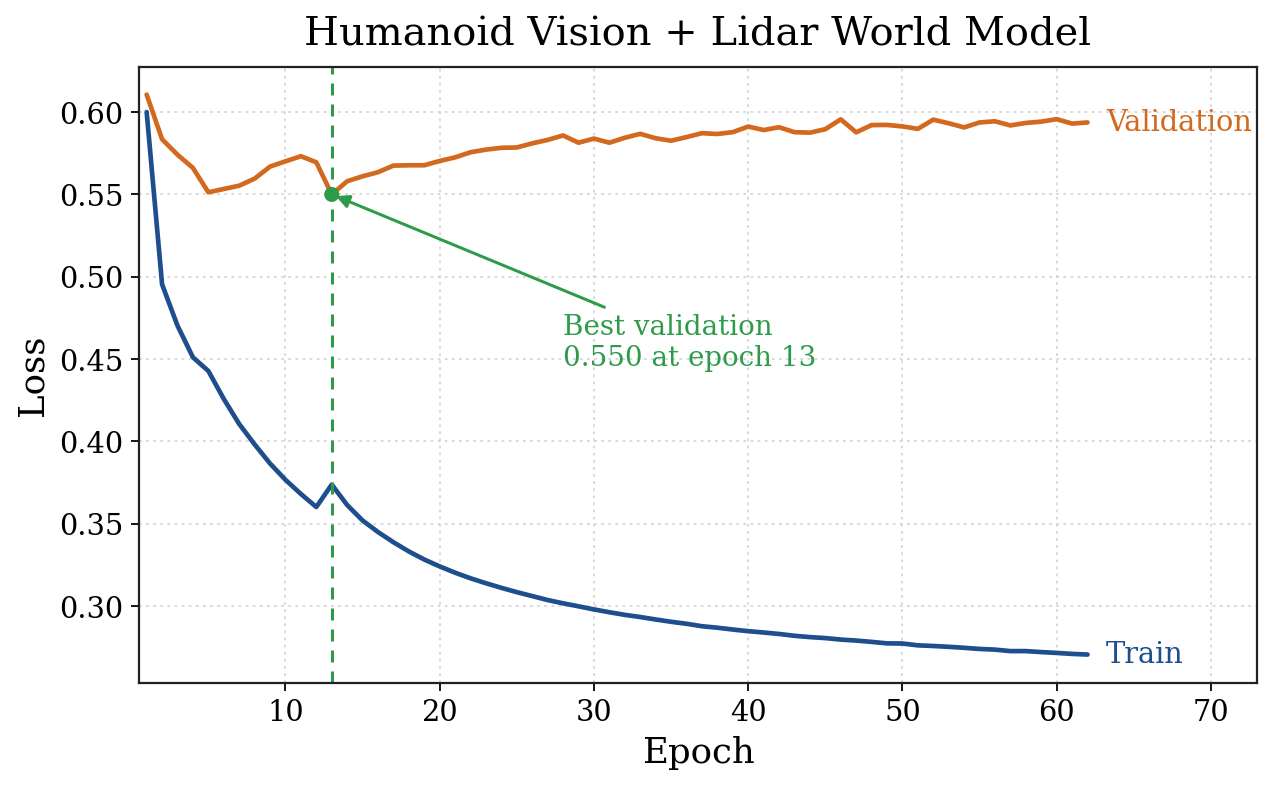}
        \par (c) Vision-LiDAR world model
    \end{minipage}
    \caption{Training and validation loss curves of the vision-based, LiDAR-based, and Vision-LiDAR world models.}
    \label{fig:wm_loss}
\end{figure}

Figure~\ref{fig:wm_loss} plots the training and validation loss of the vision-based world model during training. The training loss decreases gradually while the validation loss decreases at first and then stabilizes at about $0.7$. The lowest validation loss of $0.691$ was achieved at epoch $27$. After that, the loss gradually increases even though the training loss keeps decreasing. This indicates overfitting of the model. Thus, the checkpoint at epoch $27$ is chosen for training the safety filter.

\begin{figure*}[!htbp]
    \centering
    \includegraphics[width=\textwidth]{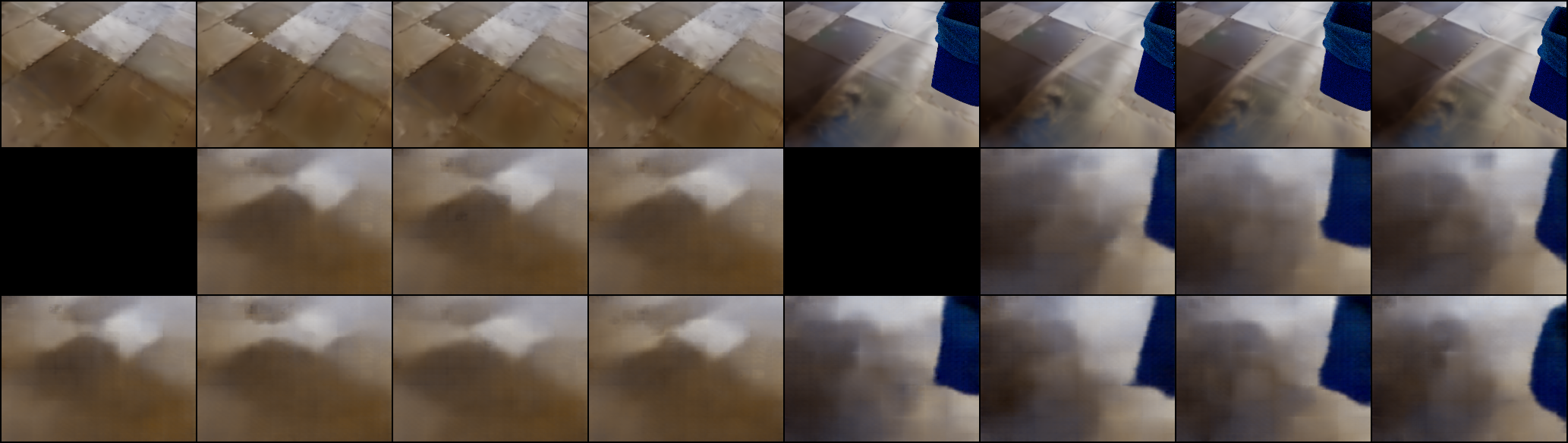}
    \caption{Qualitative reconstruction and prediction results of the vision-based world model on validation trajectories. The top row shows the original RGB observations, the middle row shows the predicted latent representations decoded into RGB images, and the bottom row shows the reconstructed images obtained by decoding the encoded visual representations.}
    \label{fig:vision_wm_prediction}
\end{figure*}

Figure~\ref{fig:vision_wm_prediction} presents a qualitative validation of the vision-based world model. The upper row shows the RGB images obtained from the simulator, the middle row shows the predicted latent states decoded to images, and the lower row shows the images that are directly reconstructed from the encoded latent states. While the images obtained from the decoder and from the direct reconstruction from the encoded states have blurred fine-grained information, they are able to capture the major structure of the environment.

\chapter{Sim-to-Real Transfer}
\label{chap:sim_to_real}

While simulation can be very useful for the development and testing of safety filters in a controlled environment, there are several errors that might occur when we deploy it in real life, such as sensor noise, delays in the control, and the difference in dynamics between robots. In this work, we transfer the previously developed safety filtering framework from simulation to a physical humanoid robot, a Unitree G1 humanoid, in order to test its real-world applicability.

\section{Real-World Robotic Platform and Control Pipeline}
\label{sec:real_robot_platform}

\noindent\textbf{Humanoid Robot Platform}

\begin{figure}[!htbp]
    \centering
    \includegraphics[width=0.65\textwidth]{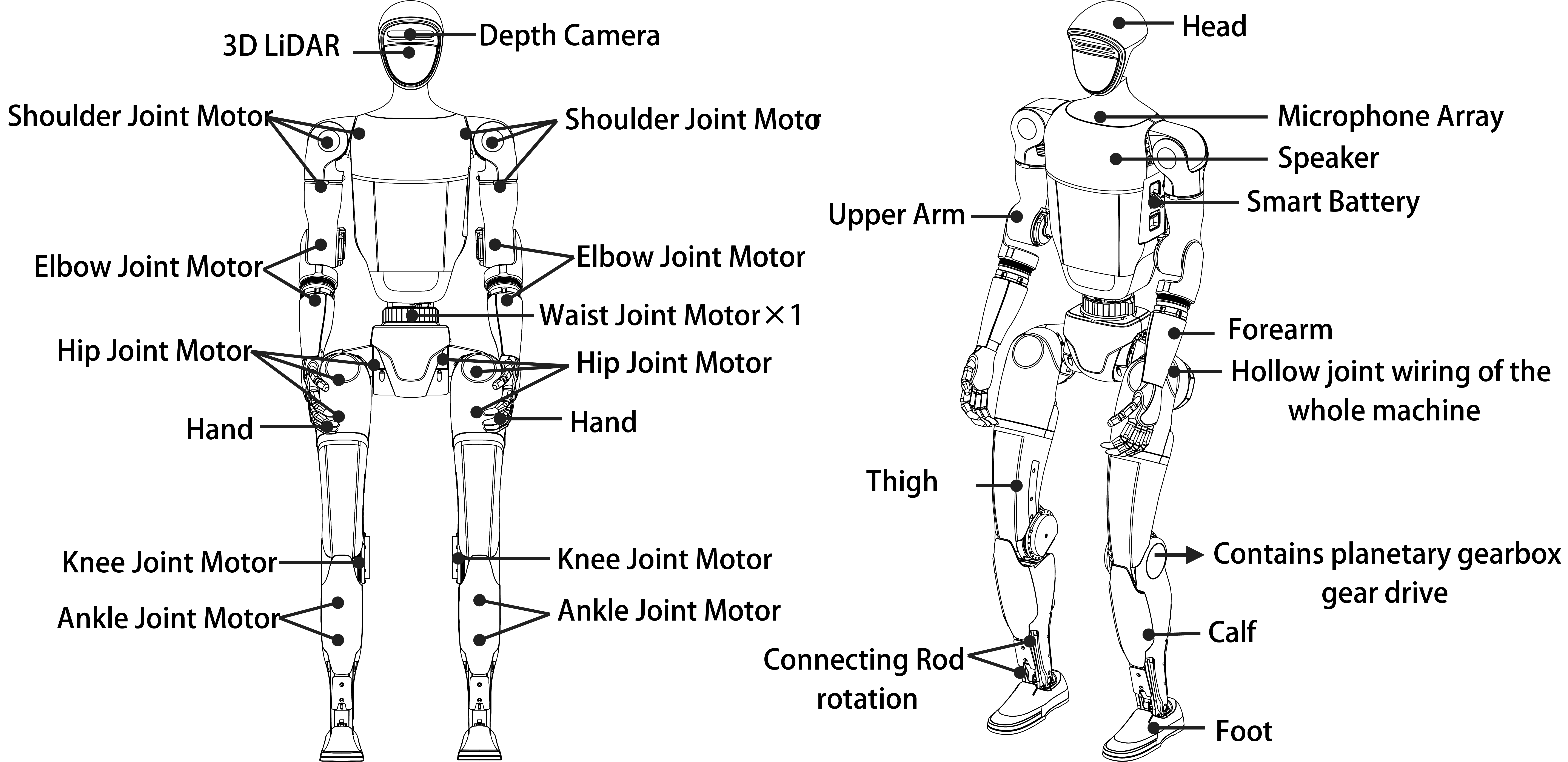}
    \caption{Hardware configuration of the Unitree G1 humanoid robot, showing its articulated structure and onboard sensing components. Source: Unitree Robotics \cite{unitreeG1Manual}.}
    \label{fig:real_robot}
\end{figure}

We conduct real-world experiments using the Unitree G1 humanoid robot, shown in Fig.~\ref{fig:real_robot}. The G1 is approximately 1.32 m tall and weighs 35 kg, with 29 degrees of freedom. Each leg has six degrees of freedom, enabling dynamic bipedal locomotion.

The Unitree G1 is controlled using the high-level locomotion interface provided by the Unitree SDK2. Similar to the simulation environment, safety filtering is performed at the high-level command layer and not on the joint level of the robot. Hence, the safety filtering framework modifies the velocity command $\mathbf{a}_{\mathrm{nom}} = (v_x, v_y, \omega)$ generated by the nominal controller for the robot in case of an unsafe situation. This command is then executed by the onboard locomotion controller of the humanoid robot.

We use the forward-facing direction as the positive $x$-axis in the robot’s local coordinate frame. For each navigation task the target waypoints are defined in terms of the robot’s initial pose. The odometry for the robot’s locomotion estimates the robot’s position and orientation as the robot moves. This information is used by the waypoint controller to estimate the waypoint’s relative displacement from the robot’s current pose. This task is always running whether the robot is executing the nominal command or the safety-filtered command. Thus, when control is returned to the nominal waypoint controller, it can recompute $\mathbf{a}_{\mathrm{nom}}$ for the robot’s current pose and continue to navigate to the waypoint that was originally specified for the task without having to redefine the waypoint.

\noindent\textbf{Onboard Sensors}

The G1 is equipped with various sensors during locomotion. The Intel RealSense RGB-D camera (15 Hz) gives us visual information of the environment. The Livox MID-360 LiDAR (7 Hz) provides us with geometric information in the form of 3D point clouds. Figure~\ref{fig:real_robot} shows where these sensors are located. This information is supplemented by the robot’s proprioception: inertial, joint position, joint velocity, and joint force information. All this information is available at around 200 Hz from the on-board IMU. By combining this information with the robot’s odometry, the robot can estimate its current locomotion state and compute corresponding waypoint-relative navigation commands.

\noindent\textbf{Computing and Communication Architecture}

\begin{figure}[!htbp]
    \centering
    \includegraphics[width=\textwidth]{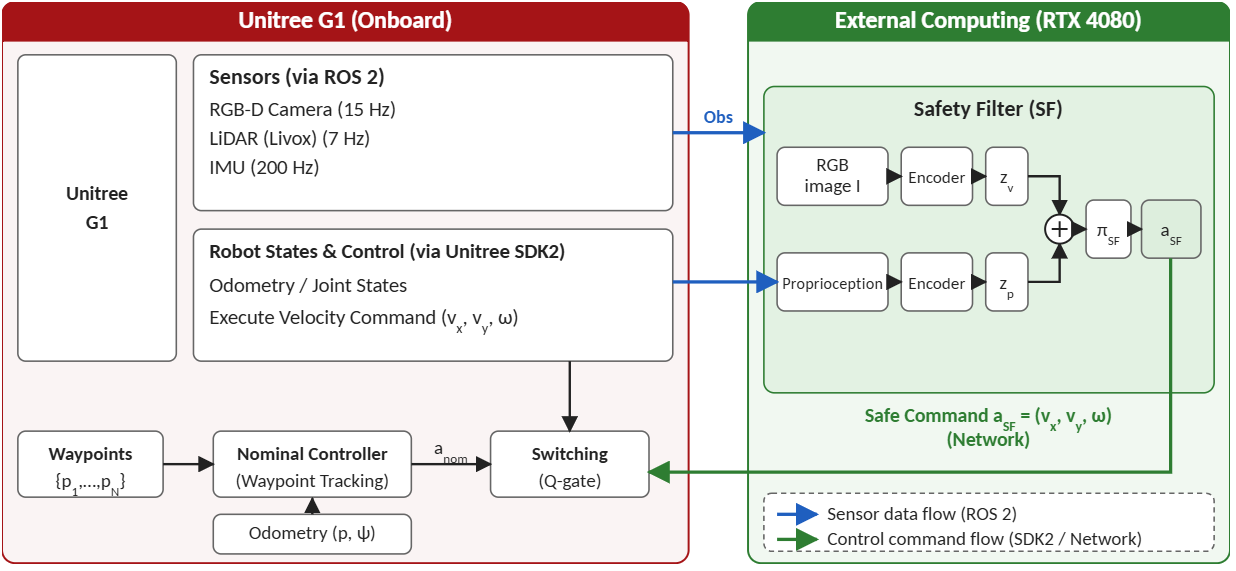}
    \caption{System architecture of the real-world safety filtering framework on the Unitree G1 humanoid robot. ROS 2 provides sensor observations, and Unitree SDK2 handles robot state estimation and locomotion control. An external NVIDIA RTX 4080 GPU performs safety filter inference, and the resulting commands are executed through the robot's onboard locomotion controller.}
    \label{fig:real_world_architecture}
\end{figure}

This real-world implementation of the computing and communication architecture is shown in Fig.~\ref{fig:real_world_architecture}. The system uses ROS 2 Foxy for collecting the RGB images, LiDAR point cloud and any other sensor observations. Unitree SDK2 is utilized for retrieving the robot’s state and high-level control of locomotion. Note that the two frameworks are in separate DDS domains. Therefore, a listener is implemented to collect and synchronize data from both communication frameworks.

Inference for the safety filter is running on an external computer which is equipped with an NVIDIA RTX 4080 GPU. The onboard computer of the robot first collects relevant observations and then computes nominal velocity commands based on our waypoint controller in Algorithm~\ref{alg:waypoint_controller}. Based on these required observations and computed commands, the external inference server then evaluates safety and either leaves commands as they are or changes them to alternative ones using our Algorithm~\ref{alg:safety_filter}. These in the end are executed via Unitree SDK2. Thus safety can be running in real time without the neural networks running on the small onboard computer of the robot to make motion smooth.

\section{Sim-to-Real Deployment}
\label{sec:sim_to_real_deployment}

\subsection{Observations and fine-tuning}
\label{subsec:real_observation_control}

\noindent\textbf{Real-World Sensor Observations}

\begin{figure}[!htbp]
    \centering
    \begin{minipage}[c]{0.43\textwidth}
        \centering
        \includegraphics[height=5cm,keepaspectratio]{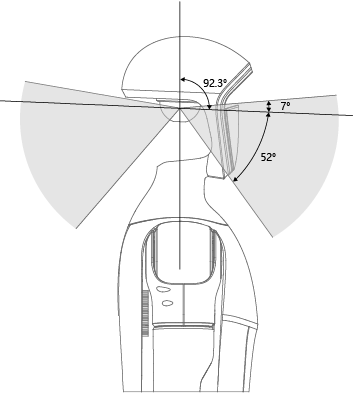}
        \par (a) Livox MID-360 LiDAR \cite{livoxMid360}
    \end{minipage}
    \hfill
    \begin{minipage}[c]{0.43\textwidth}
        \centering
        \includegraphics[height=5cm,keepaspectratio]{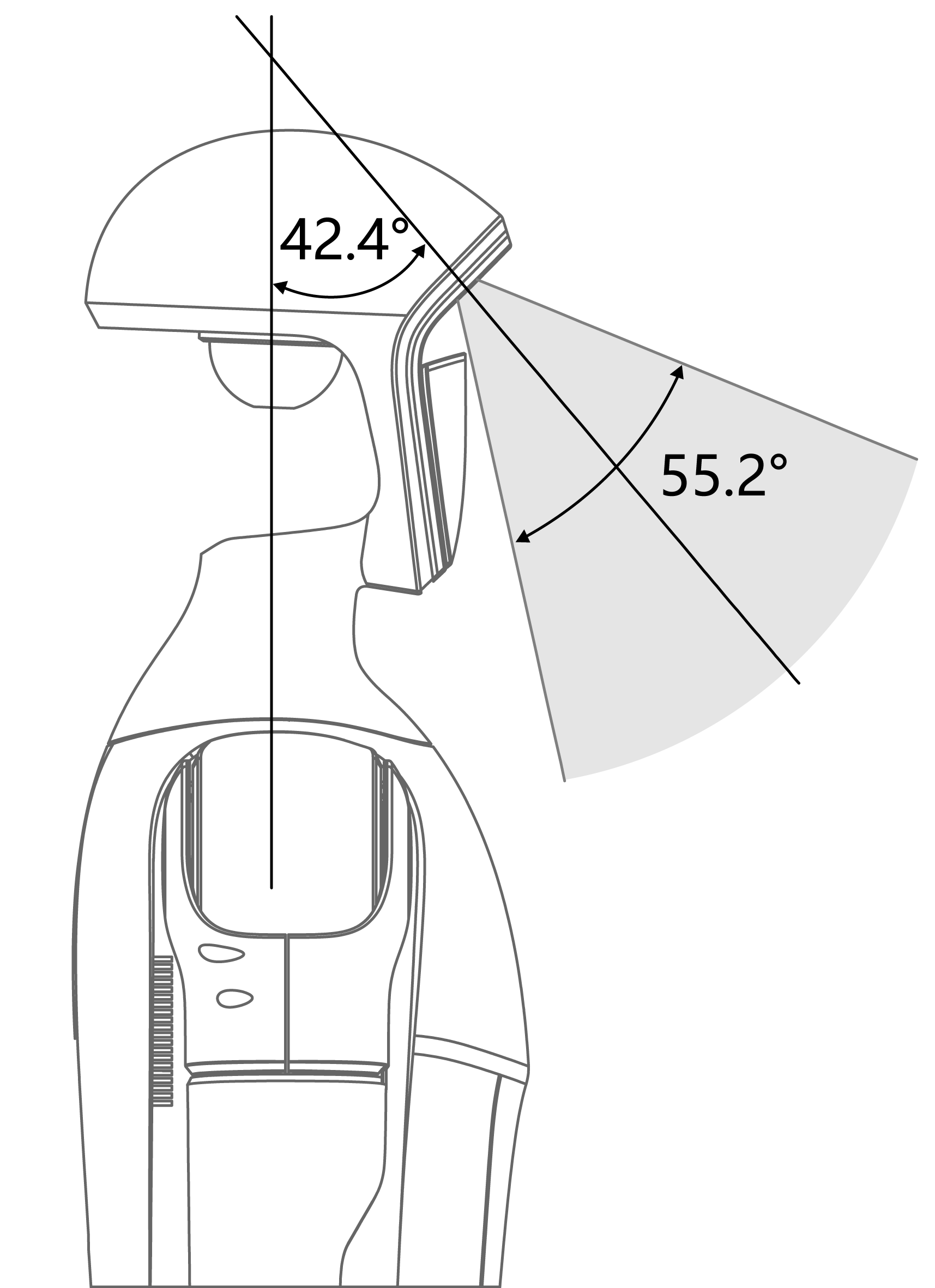}
        \par (b) Intel RealSense D435i \cite{realsenseD435i}
    \end{minipage}
    \caption{Field-of-view configurations of the real-world perception sensors: (a) Livox MID-360 LiDAR and (b) Intel RealSense D435i RGB-D camera.}
    \label{fig:real_sensor_fov}
\end{figure}

The real-world safety filtering system is composed of an Intel RealSense D435i RGB-D camera and a Livox MID-360 LiDAR for environmental perception in Fig.~\ref{fig:real_sensor_fov}. The downward tilt angle of the D435i’s optics is about $42.4^\circ$. It acquires forward-facing color/depth images at 15 Hz. The MID-360 LiDAR has a $360^\circ$ horizontal field of view and a $59^\circ$ vertical field of view. Its effective sensing area is between $7^\circ$ above the horizontal plane and $52^\circ$ below the horizontal plane. It acquires point cloud data at about 7 Hz.

\noindent\textbf{Observation Preprocessing}

For the vision input we resize the captured RGB images from the RealSense camera to $192 \times 256$ pixels using bilinear interpolation. These images are then sent to the inference server. After that the images are normalized to the interval [-1,1] and then processed by the frozen DINOv2 encoder. In the meantime the joint positions of the robot are padded from 29 to 37 dimensions in order to match the corresponding simulation configuration. Then the generated visual and proprioceptive features are concatenated and after a temporal averaging over the last 10 frames the resulting feature is used as a latent state by the safety filter.

We process raw point cloud data from the Livox MID-360 LiDAR. First, we remove ground points, points that reflect off the robot body and points that are farther than 6 m from the sensor. The points are transformed into the simulated LiDAR sensor coordinate frame. We then voxelize the remaining points at a resolution of 2 cm and pass them through the frozen SpUNet encoder to a 96-dimensional LiDAR feature space. This feature space is then concatenated with proprioceptive features. The resulting 106-dimensional latent space is then temporally averaged over the last 10 frames and passed through the safety filter.

\subsection{Fine-Tuning Strategy}
\label{subsec:real_fine_tuning}

To better transfer the safety to real-world settings, we fine-tune the safety filters that were pre-trained in the simulated environment with real-world data collected from a Unitree G1 robot. 157 trajectories were manually collected using a remote controller for the robot. We recorded sensor measurements, the commanded velocities, and the distance to the closest obstacle measured by LiDAR for all episodes. The collected data was saved at 15 Hz. We fine-tuned the safety filters in a completely offline fashion using these trajectories and no further interaction with the physical robot or the simulation environment was required.

\noindent\textbf{Real-World Experience Replay}

These recorded RGB images and LiDAR point clouds in episodes are processed in separate observation pipelines as for the simulations above. All of the visual and LiDAR encoders and the proprioceptive encoder are then frozen for fine-tuning of the three networks above, and each observation in an episode is mapped to a latent state $\mathbf{z}_t$ where the same temporal averaging as for the simulations is applied to these. The recorded velocity actions $\mathbf{a}_t$ are constrained to the same action bounds as used in the simulations above. An episode is mapped to a sequence of transitions of the following form

\begin{equation}
(\mathbf{z}_t,\mathbf{a}_t,h_s(t+1),\mathbf{z}_{t+1}),
\end{equation}
where the safety signal is computed from the obstacle distance as follows:

\begin{equation}
h_s(t+1)=d_{t+1}-d_{\mathrm{safe}},
\qquad d_{\mathrm{safe}}=1.0~\mathrm{m}.
\end{equation}
Note that a negative safety signal means the robot has entered a predefined unsafe area. Transition are stored in the replay buffer for offline training.

\noindent\textbf{Safety Filter Fine-Tuning}

The actor and twin critics are initialized with the corresponding avoid-SAC models trained in simulation in Chapter~\ref{chap:latent_based_safety_filtering}. For both the vision-based and LiDAR-based safety filters, the actor and two critics are jointly fine-tuned using real-world data under the same avoid-SAC objective, with a discount factor of $\gamma=0.98$. The target critic networks are updated through soft updates with $\tau=0.005$. Fine-tuning uses mini-batches of 512 transitions sampled from the real-world replay buffer, with 250 gradient updates per epoch over 50 epochs. The resulting actor and critic parameters are used for subsequent real-world safety filtering experiments.

\section{Real-World Experiments}
\label{sec:real_world_experiments}

To evaluate the safety filtering framework beyond simulation, we conducted real-world experiments using the Unitree G1 humanoid robot. This section describes the experimental setup and evaluates the performance of the safety filters in real-world obstacle avoidance tasks.

\subsection{Experimental Setup}
\label{subsec:real_experimental_setup}

We evaluated the safety filters before and after fine-tuning using a blue bin as an obstacle in our laboratory environment. To ensure a fair and reproducible comparison, we randomly generated 10 configurations, each specifying the positions of the robot's starting point, goal, and obstacle. These configurations were fixed and reused across all safety filtering methods to ensure that each method was evaluated under identical conditions. In each configuration, the robot's starting position and goal were aligned along a straight path, with the obstacle placed directly in front of the robot close to this path. Therefore, the nominal controller without safety filtering was expected to collide with the obstacle.

We evaluated performance using the same five metrics as in Section~\ref{subsec:evaluation_metrics}: success rate, collision rate, average velocity, filter activation rate, and minimum obstacle distance, consistent with simulation experiments. These metrics measure the ability of each safety filter to complete navigation tasks, avoid collisions, maintain locomotion efficiency, and intervene in nominal control when necessary.

\begin{figure}[!htbp]
    \centering
    \begin{minipage}[t]{0.48\textwidth}
        \centering
        \includegraphics[width=\linewidth,height=5cm,keepaspectratio]{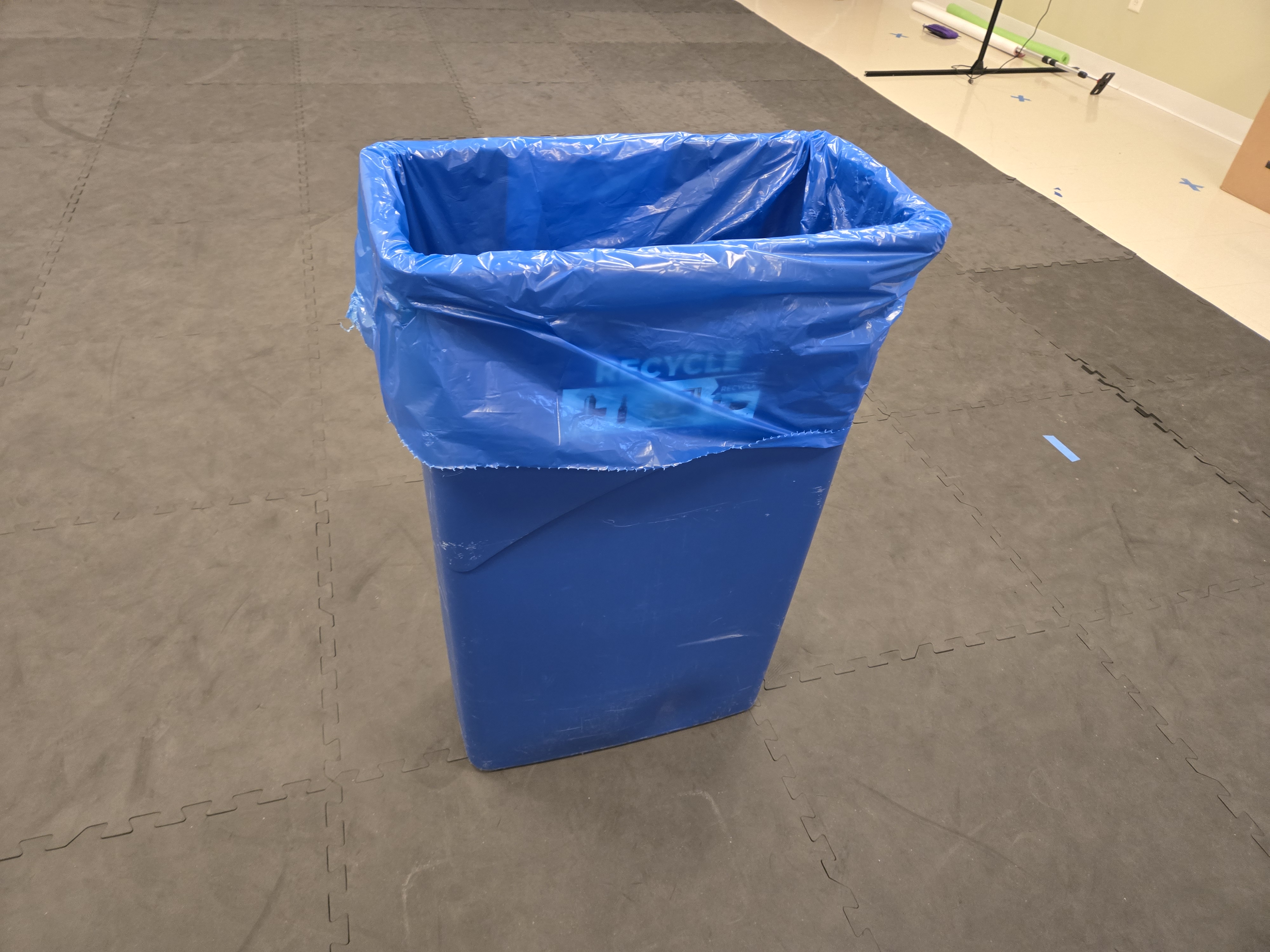}
        \par (a) Blue bin obstacle
    \end{minipage}
    \hfill
    \begin{minipage}[t]{0.48\textwidth}
        \centering
        \includegraphics[width=\linewidth,height=5cm,keepaspectratio]{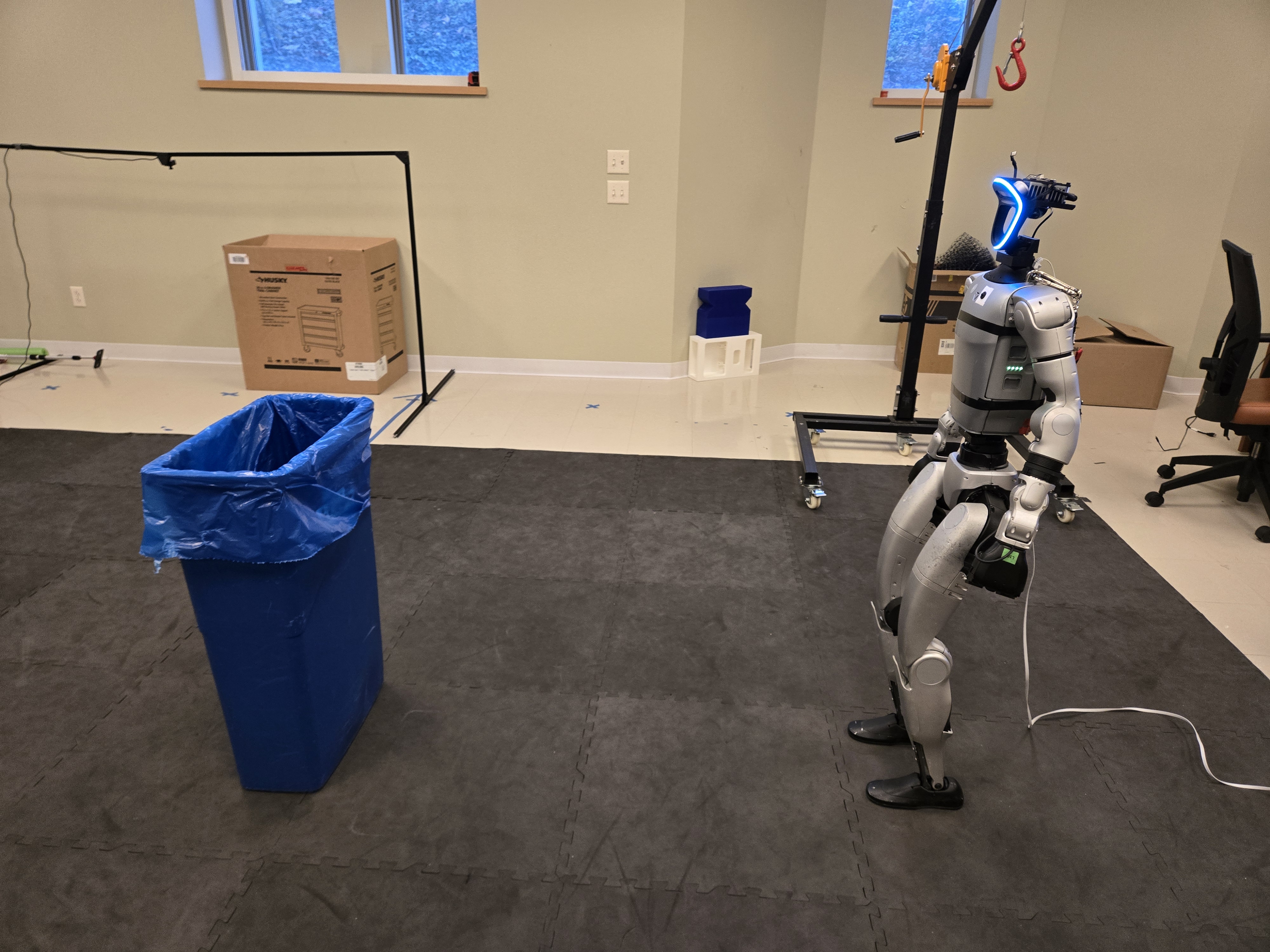}
        \par (b) Unitree G1 facing the obstacle
    \end{minipage}
    \caption{Real-world experimental setup. (a) Blue bin used as an obstacle. (b) Unitree G1 humanoid robot positioned in front of the obstacle during evaluation.}
    \label{fig:real_world_setup}
\end{figure}

\subsection{Safety Filtering Performance}
\label{subsec:real_safety_performance}

To evaluate the six safety filters we conducted two different experiments: one before fine-tuning on real-world data and one after fine-tuning on real-world data. The results of the experiments before fine-tuning are shown in Table~\ref{tab:real_safety_filter_before_finetuning}, and the results of the experiments after fine-tuning are shown in Table~\ref{tab:real_safety_filter_comparison}. We measure the results using the five metrics. The success rate and the collision rate of a method measure if the method is able to complete a task safely. The average velocity of a method measures how fast the method can move. The filter activation rate of a method measures how often the safety filter of a method interferes with the nominal controller of the method. The minimum obstacle distance of a method measures how close a robot of a method gets to an obstacle while the method is completing a task.

\begin{table}[!htbp]
\centering
\caption{Comparison of safety filtering methods before fine-tuning in real-world experiments}
\label{tab:real_safety_filter_before_finetuning}
\resizebox{\textwidth}{!}{%
\begin{tabular}{lccccc}
\hline
\textbf{Filter} &
\shortstack{\textbf{Success}\\\textbf{Rate} $\uparrow$} &
\shortstack{\textbf{Collision}\\\textbf{Rate} $\downarrow$} &
\shortstack{\textbf{Average}\\\textbf{Velocity (m/s)} $\uparrow$} &
\shortstack{\textbf{Filter Activation}\\\textbf{Rate} $\downarrow$} &
\shortstack{\textbf{Minimum Obstacle}\\\textbf{Distance (m)} $\uparrow$} \\
\hline
No Filter & 0 & 1 & N/A & $\mathbf{0.000 \pm 0.000}$ & $0.000 \pm 0.000$ \\
Lidar-only SF & 0.5 & 0.3 & $0.219 \pm 0.051$ & $0.453 \pm 0.268$ & $0.532 \pm 0.224$ \\
Vision-only SF & 0.2 & 0.8 & $0.205 \pm 0.026$ & $0.487 \pm 0.329$ & $0.412 \pm 0.196$ \\
Lidar-Vision SF & $\mathbf{0.6}$ & $\mathbf{0.2}$ & $0.237 \pm 0.039$ & $0.382 \pm 0.241$ & $0.594 \pm 0.213$ \\
Lidar-only WM SF & 0.5 & $\mathbf{0.2}$ & $0.220 \pm 0.046$ & $0.472 \pm 0.276$ & $0.548 \pm 0.228$ \\
Vision-only WM SF & 0.2 & 0.8 & $0.211 \pm 0.031$ & $0.468 \pm 0.312$ & $0.443 \pm 0.207$ \\
Lidar-Vision WM SF & $\mathbf{0.6}$ & $\mathbf{0.2}$ & $\mathbf{0.241 \pm 0.040}$ & $0.397 \pm 0.239$ & $\mathbf{0.601 \pm 0.215}$ \\
\hline
\end{tabular}%
}
\end{table}

Table~\ref{tab:real_safety_filter_before_finetuning} presents the performance of the six safety filtering methods before fine-tuning. Compared to the nominal controller without safety filtering, all six methods showed improvements in obstacle avoidance and task completion. The best performance was achieved by the two methods combining visual and LiDAR information, LiDAR-Vision SF and LiDAR-Vision WM SF, both reaching a success rate of 0.6 and a collision rate of 0.2. These shows that combining both sensing modalities provides more reliable obstacle avoidance than relying on either modality alone before fine-tuning.

Among the single-modality methods, the LiDAR-only SF achieved a success rate of 0.5 and a collision rate of 0.3, while the Vision-only SF achieved a substantially lower success rate of 0.2 and a higher collision rate of 0.8. These results suggest that vision-only safety filtering is particularly vulnerable to failures before fine-tuning. The impact of incorporating world models also varied between sensing modalities. For Vision-only SF, the addition of a world model did not change the success or collision rates, while for LiDAR-only SF, the collision rate decreased from 0.3 to 0.2 and the success rate remained unchanged. In contrast, LiDAR-Vision WM SF achieved the same success and collision rates as LiDAR-Vision SF, at 0.6 and 0.2, respectively. Overall, the multimodal safety filtering methods demonstrated the best balance between task completion, collision avoidance, and movement efficiency before fine-tuning. However, adding a world model provided only limited performance improvements across all safety filtering methods.

\begin{table}[!htbp]
\centering
\caption{Comparison of fine-tuned safety filtering methods in real-world experiments}
\label{tab:real_safety_filter_comparison}
\resizebox{\textwidth}{!}{%
\begin{tabular}{lccccc}
\hline
\textbf{Filter} &
\shortstack{\textbf{Success}\\\textbf{Rate} $\uparrow$} &
\shortstack{\textbf{Collision}\\\textbf{Rate} $\downarrow$} &
\shortstack{\textbf{Average}\\\textbf{Velocity (m/s)} $\uparrow$} &
\shortstack{\textbf{Filter Activation}\\\textbf{Rate} $\downarrow$} &
\shortstack{\textbf{Minimum Obstacle}\\\textbf{Distance (m)} $\uparrow$} \\
\hline
No Filter & 0 & 1 & N/A & $\mathbf{0.000 \pm 0.000}$ & $0.000 \pm 0.000$ \\
Lidar-only SF & 0.7 & 0.1 & $0.241 \pm 0.042$ & $0.391 \pm 0.247$ & $0.621 \pm 0.205$ \\
Vision-only SF & 0.4 & 0.5 & $0.227 \pm 0.018$ & $0.425 \pm 0.313$ & $0.489 \pm 0.180$ \\
Lidar-Vision SF & \textbf{0.8} & 0.1 & $0.255 \pm 0.031$ & $0.315 \pm 0.205$ & $\mathbf{0.680 \pm 0.195}$ \\
Lidar-only WM SF & 0.7 & 0.1 & $0.238 \pm 0.039$ & $0.410 \pm 0.255$ & $0.610 \pm 0.210$ \\
Vision-only WM SF & 0.5 & 0.4 & $0.235 \pm 0.025$ & $0.400 \pm 0.290$ & $0.510 \pm 0.190$ \\
Lidar-Vision WM SF & \textbf{0.8} & 0.1 & $\mathbf{0.260 \pm 0.034}$ & $0.330 \pm 0.220$ & $0.665 \pm 0.200$ \\
\hline
\end{tabular}%
}
\end{table}

Table~\ref{tab:real_safety_filter_comparison} shows the performance of the six safety filtering methods after fine-tuning with real robot data. All safety filtering methods improved the robot's ability to avoid obstacles compared to the nominal controller without safety filtering. The best results were still achieved by the two methods that combine information from visual information and LiDAR data, LiDAR-Vision SF and LiDAR-Vision WM SF, reaching a success rate of 0.8 while keeping the collision rate at 0.1. Thus, the combination of both sensing modalities outperforms the individual sensing modalities.

We compared the performance of the three different safety filtering methods without world models. The LiDAR-only SF achieved a success rate of 0.7 and a collision rate of 0.1, while the Vision-only SF achieved a success rate of 0.4 and a collision rate of 0.5. These results show that the LiDAR-only SF has a smaller sim-to-real gap than the Vision-only SF. After fine-tuning with real-world data, the LiDAR-only SF achieved performance comparable to its simulation results reported in Table~\ref{tab:safety_filter_comparison}, whereas the Vision-only SF still performed substantially worse than in simulation. The LiDAR-Vision SF achieved a success rate of 0.8, a collision rate of 0.1, and the largest minimum obstacle distance of $0.680$ m among the three safety filtering methods. The minimum obstacle distances of LiDAR-only SF and Vision-only SF were $0.621$ m and $0.489$ m, respectively. These results demonstrate that the combination of LiDAR and vision achieved the highest success rate, the lowest collision rate, and the best obstacle clearance among the three safety filtering methods in real-world environments.

Our results for the impact of world models included in safety filtering vary with sensing modality. For Vision-only SF, the addition of a world model raises the success rate to 0.5 while lowering the collision rate to 0.4. For the LiDAR-only SF, the success and collision rates for safety filtering remain unchanged at 0.7 and 0.1, respectively. For the LiDAR-Vision WM SF, the success and collision rates for safety filtering are also unchanged from the LiDAR-Vision SF at 0.8 and 0.1, respectively. Thus, while there is some benefit from inclusion of a world model with Vision-only safety filtering, the benefits of including a world model with safety filtering that utilizes LiDAR information are not significant.

For the movement efficiency, the LiDAR-Vision WM SF reached the highest average velocity of $0.260$ m/s. This was followed by the LiDAR-Vision SF with an average velocity of $0.255$ m/s. In addition, the LiDAR-Vision SF had the lowest filter activation rate of all six methods with $0.315$. This shows that the LiDAR-Vision SF required less frequent intervention in the nominal controller in order to reach high task success and to safely clear obstacles. In the end, the multimodal safety filtering was able to achieve the best trade-off between completing the task, avoiding collisions, and moving efficiently in the real-world experiments.

\begin{figure}[H]
    \centering
    \includegraphics[width=0.48\textwidth]{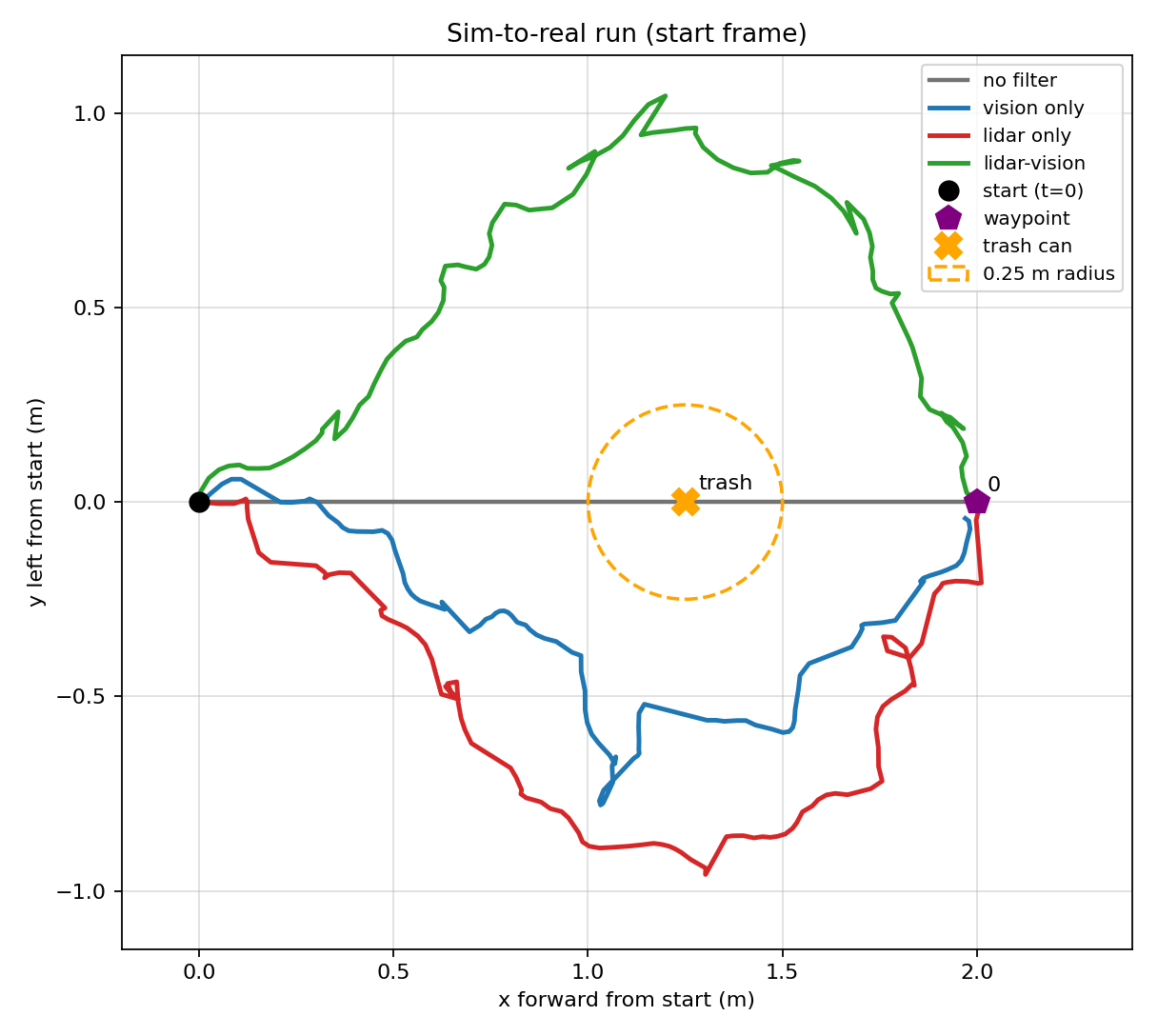}
    \caption{Top-down visualization of real-world obstacle avoidance trajectories of the Unitree G1 humanoid robot using the Vision-only SF, LiDAR-only SF, LiDAR-Vision SF, and nominal controller, with a blue bin of approximately 0.25 m in radius as the obstacle.}
    \label{fig:real_trajectory_comparison}
\end{figure}

Figure~\ref{fig:real_trajectory_comparison} shows a top-down view of the real-world locomotion trajectories of the Unitree G1 humanoid robot using different safety filtering methods. The robot starts at the origin and navigates toward a waypoint located 2 m ahead, with an obstacle placed along the nominal path. Without safety filtering, the robot follows a straight path toward the obstacle. After applying the three safety filter methods, all methods modify the nominal velocity commands to navigate around the obstacle and reach the target waypoint. The Vision-only SF and LiDAR-only SF guide the robot around the right side of the obstacle, with the LiDAR-only SF maintaining a greater distance from the obstacle, while the LiDAR-Vision SF guides the robot around the left side.

\begin{figure}[H]
    \centering
    \begin{minipage}[t]{0.49\textwidth}
        \centering
        \includegraphics[width=\linewidth]{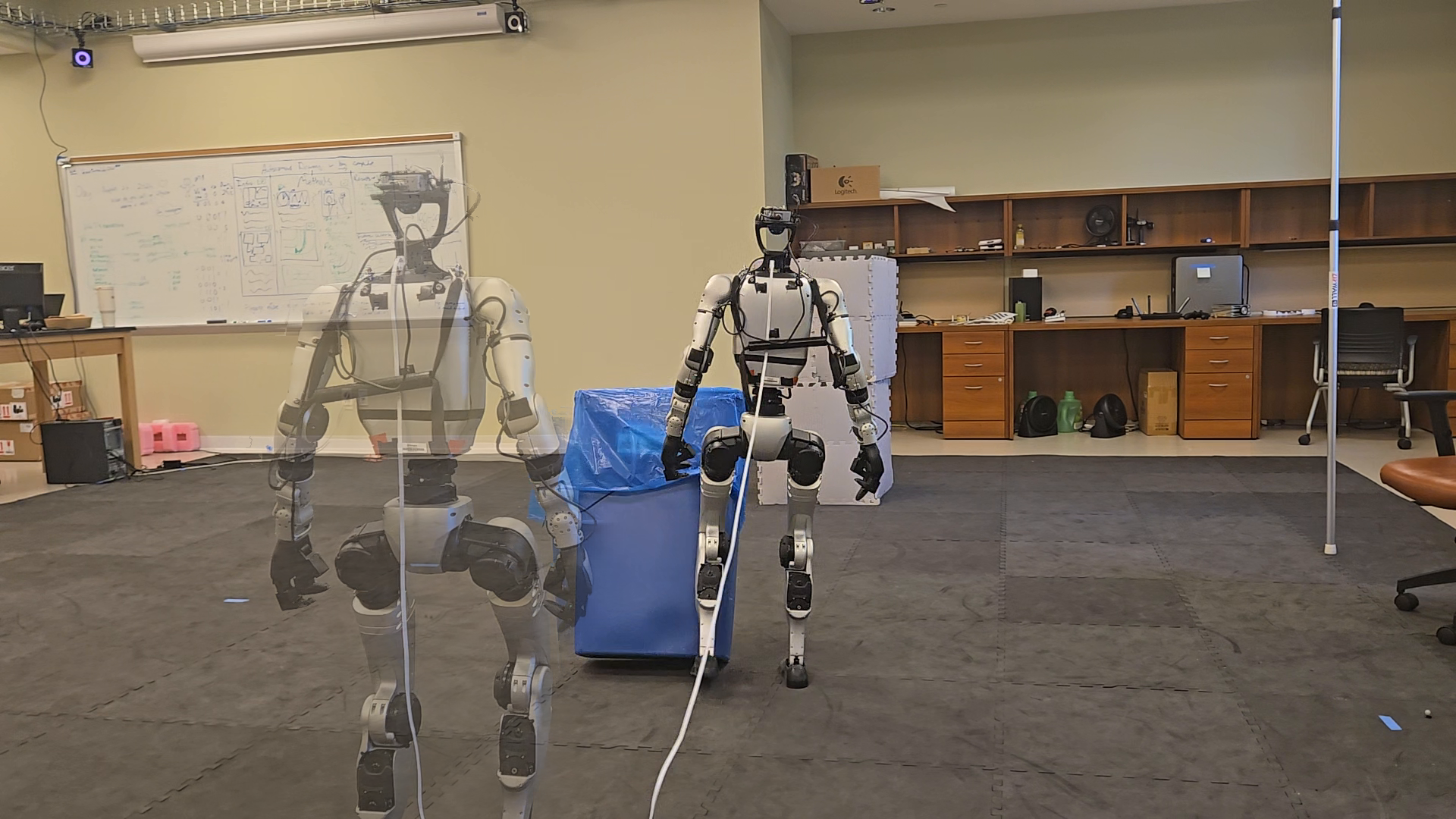}
        \small (a) Before fine-tuning
    \end{minipage}
    \hfill
    \begin{minipage}[t]{0.49\textwidth}
        \centering
        \includegraphics[width=\linewidth]{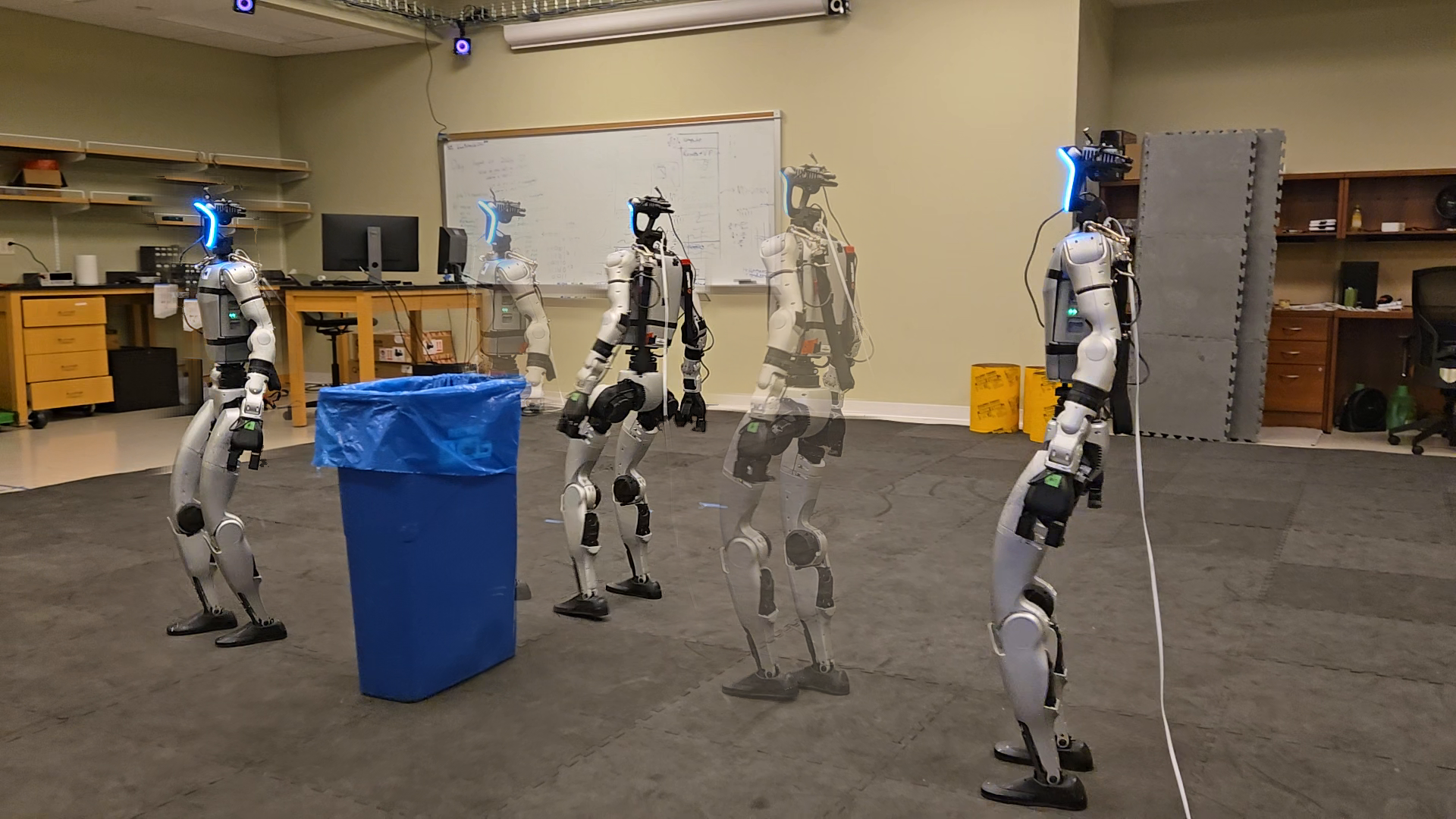}
        \small (b) After fine-tuning
    \end{minipage}
    \caption{Qualitative visualization of real-world obstacle avoidance using the vision-only safety filter on the Unitree G1 humanoid robot: (a) before fine-tuning and (b) after fine-tuning. Robot poses at different time steps are overlaid on a single image to illustrate the obstacle avoidance trajectories.}
    \label{fig:real_obstacle_avoidance}
\end{figure}

Figure~\ref{fig:real_obstacle_avoidance} provides a third-person visualization of the robot's obstacle avoidance behavior using the Vision-only SF before and after fine-tuning. Before fine-tuning, the robot failed to avoid the obstacle, whereas after fine-tuning, the robot successfully navigated around the obstacle and reached its target.

\chapter{Conclusions and Future Work}
\label{chap:conclusion and future work}

In this work, we investigate latent-based safety filtering for humanoid robot locomotion by developing and evaluating safety filters using LiDAR, vision, and their combination on the Unitree G1 humanoid robot. We compare six safety filtering methods, including three sensing configurations with and without world models, in both simulation and real-world experiments. Our results demonstrate that combining LiDAR and vision provides more reliable obstacle avoidance than relying on either sensing modality alone. In real-world experiments, the fine-tuned LiDAR-Vision safety filters achieved a success rate of 0.8 and a collision rate of 0.1, outperforming the single-modality methods. We also investigate the impact of world models and real-world fine-tuning, showing that fine-tuning improves safety and movement efficiency, while the benefits of incorporating world models vary across sensing modalities. Overall, our results demonstrate the effectiveness of multimodal perception for improving the safety and reliability of humanoid locomotion.

Our future plans are: (1) developing a reachability-based multimodal fusion method that combines separately learned LiDAR and vision value functions through their minimum to train a more conservative safety actor, (2) incorporating obstacle bounding boxes and geometric features extracted from images to improve the generalization of vision-based safety filtering to different obstacle types, and (3) extending world model predictions beyond the next latent state to capture longer-term environmental information for safer decision-making.


\bibliographystyle{abbrv}
\begin{spacing}{1.0}
\bibliography{thesis-references}

@article{min2026autonomous,
author = {Min, Chen and Si, Shubin and Wang, Xu and Xue, Hanzhang and Jiang, Weizhong and Chen, Zitong and Li, Mengmeng and Mei, Jilin and Shang, Erke and Xiao, Zhipeng and Dai, Bin and Zhu, Qi and Fu, Hao and Zhao, Dawei and Xiao, Liang and Nie, Yiming and Hu, Yu},
title = {Autonomous Driving in Unstructured Off-Road Environments: How Far Have We Come?},
journal = {Journal of Field Robotics},
volume = {43},
number = {5},
pages = {3403-3439},
doi = {https://doi.org/10.1002/rob.70154},
url = {https://onlinelibrary.wiley.com/doi/abs/10.1002/rob.70154},
eprint = {https://onlinelibrary.wiley.com/doi/pdf/10.1002/rob.70154},
year = {2026}
}

@ARTICLE{HumanoidRobotLocomotionSurveyandPerspectives,
  author={Cao, Yifeng and He, Junpeng and Li, Bingxian and Xu, Songchen and Tao, Xiaowen and Fan, Lili and Wen, Weisong and Cao, Dongpu},
  journal={IEEE Robotics \& Automation Magazine}, 
  title={Humanoid Robot Locomotion in Unstructured Environments: A Survey and Perspectives}, 
  year={2026},
  volume={},
  number={},
  pages={2-15},
  doi={10.1109/MRA.2026.3703697},
  ISSN={1558-223X},
  month={},}

@article{ha2025legged,
author = {Sehoon Ha and Joonho Lee and Michiel van de Panne and Zhaoming Xie and Wenhao Yu and Majid Khadiv},
title ={Learning-based legged locomotion: State of the art and future perspectives},

journal = {The International Journal of Robotics Research},
volume = {44},
number = {8},
pages = {1396-1427},
year = {2025},
doi = {10.1177/02783649241312698},

URL = { 
    
        https://doi.org/10.1177/02783649241312698
    
    

},
eprint = { 
    
        https://doi.org/10.1177/02783649241312698
    
    

}
}

@article{Freyleggedrobots,
author = {Jonas Frey  and Matías Mattamala  and Hae-Won Park  and Mayank Mittal  and Georg Martius  and Maike Osborne  and Robert Sparrow  and Marco Hutter },
title = {Advances, challenges, and opportunities for legged robots},
journal = {Science Robotics},
volume = {11},
number = {116},
pages = {eaee0787},
year = {2026},
doi = {10.1126/scirobotics.aee0787},
URL = {https://www.science.org/doi/abs/10.1126/scirobotics.aee0787},
eprint = {https://www.science.org/doi/pdf/10.1126/scirobotics.aee0787}}

@ARTICLE{KurniawanLeggedRobotLocomotion,
  author={Kurniawan, Irfan Tito and Zhu, Wei and Owaki, Dai and Hayashibe, Mitsuhiro},
  journal={IEEE Robotics \& Automation Magazine}, 
  title={Learning Perceptive Legged Robot Locomotion in the Real World: A Systematic Review}, 
  year={2026},
  volume={33},
  number={1},
  pages={8-23},
  doi={10.1109/MRA.2025.3639794},
  ISSN={1558-223X},
  month={March},}

@INPROCEEDINGS{RenVB-Com,
  author={Ren, Junli and Huang, Tao and Wang, Huayi and Wang, Zirui and Ben, Qingwei and Long, Junfeng and Yang, Yanchao and Pang, Jiangmiao and Luo, Ping},
  booktitle={2026 IEEE International Conference on Robotics and Automation (ICRA)}, 
  title={VB-Com: Learning Vision-Blind Composite Humanoid Locomotion Against Deficient Perception}, 
  year={2026},
  volume={},
  number={},
  pages={13874-13881},
  doi={10.1109/ICRA57385.2026.11695836},
  ISSN={},
  month={June},}

@InProceedings{Omni-Perception,
  title = 	 {Omni-Perception: Omnidirectional Collision Avoidance of Legged Robots in Dynamic Environments},
  author =       {Wang, Zifan and Ma, Teli and Jia, Yufei and Yang, Xun and Zhou, Jiaming and OUYANG, Wenlong and Zhang, Qiang and Liang, Junwei},
  booktitle = 	 {Proceedings of The 9th Conference on Robot Learning},
  pages = 	 {670--685},
  year = 	 {2025},
  editor = 	 {Lim, Joseph and Song, Shuran and Park, Hae-Won},
  volume = 	 {305},
  series = 	 {Proceedings of Machine Learning Research},
  month = 	 {27--30 Sep},
  publisher =    {PMLR},
  url = 	 {https://proceedings.mlr.press/v305/wang25b.html}
}

@INPROCEEDINGS{SHIELD,
  author={Yang, Lizhi and Werner, Blake and Cosner, Ryan K. and Fridovich-Keil, David and Culbertson, Preston and Ames, Aaron D.},
  booktitle={2025 IEEE/RSJ International Conference on Intelligent Robots and Systems (IROS)}, 
  title={SHIELD: Safety on Humanoids via CBFs In Expectation on Learned Dynamics}, 
  year={2025},
  volume={},
  number={},
  pages={203-210},
  doi={10.1109/IROS60139.2025.11247065},
  ISSN={2153-0866},
  month={Oct},}

@INPROCEEDINGS{PengSafeBipedalRobotNav,
  author={Peng, Chengyang and Paredes, Victor and Castillo, Guillermo A. and Hereid, Ayonga},
  booktitle={2025 IEEE International Conference on Robotics and Automation (ICRA)}, 
  title={Real-Time Safe Bipedal Robot Navigation using Linear Discrete Control Barrier Functions}, 
  year={2025},
  volume={},
  number={},
  pages={14903-14909},
  doi={10.1109/ICRA55743.2025.11128366},
  ISSN={},
  month={May},}

@ARTICLE{OneFilter,
  author={Lin, Albert and Peng, Shuang and Bansal, Somil},
  journal={IEEE Transactions on Robotics}, 
  title={One Filter to Deploy Them All: Robust Safety for Quadrupedal Navigation in Unknown Environments}, 
  year={2026},
  volume={42},
  number={},
  pages={545-560},
  doi={10.1109/TRO.2025.3644957},
  ISSN={1941-0468},
  month={},}

@article{HsuSafetyFilterReview,
author = {Hsu, Kai-Chieh and Hu, Haimin and Fisac, Jaime},
year = {2024},
month = {07},
pages = {47-72},
title = {The Safety Filter: A Unified View of Safety-Critical Control in Autonomous Systems},
volume = {7},
journal = {Annual Review of Control, Robotics, and Autonomous Systems},
doi = {10.1146/annurev-control-071723-102940}
}

@ARTICLE{AmesCBF,
  author={Ames, Aaron D. and Xu, Xiangru and Grizzle, Jessy W. and Tabuada, Paulo},
  journal={IEEE Transactions on Automatic Control}, 
  title={Control Barrier Function Based Quadratic Programs for Safety Critical Systems}, 
  year={2017},
  volume={62},
  number={8},
  pages={3861-3876},
  doi={10.1109/TAC.2016.2638961},
  ISSN={1558-2523},
  month={Aug},}

@article{COHENCBF,
title = {Safety-critical control for autonomous systems: Control barrier functions via reduced-order models},
journal = {Annual Reviews in Control},
volume = {57},
pages = {100947},
year = {2024},
issn = {1367-5788},
doi = {https://doi.org/10.1016/j.arcontrol.2024.100947},
url = {https://www.sciencedirect.com/science/article/pii/S1367578824000166},
author = {Max H. Cohen and Tamas G. Molnar and Aaron D. Ames}
}

@article{GARGCBF,
title = {Advances in the Theory of Control Barrier Functions: Addressing practical challenges in safe control synthesis for autonomous and robotic systems},
journal = {Annual Reviews in Control},
volume = {57},
pages = {100945},
year = {2024},
issn = {1367-5788},
doi = {https://doi.org/10.1016/j.arcontrol.2024.100945},
url = {https://www.sciencedirect.com/science/article/pii/S1367578824000142},
author = {Kunal Garg and James Usevitch and Joseph Breeden and Mitchell Black and Devansh Agrawal and Hardik Parwana and Dimitra Panagou}
}

@INPROCEEDINGS{BansalHJReachability,
  author={Bansal, Somil and Chen, Mo and Herbert, Sylvia and Tomlin, Claire J.},
  booktitle={2017 IEEE 56th Annual Conference on Decision and Control (CDC)}, 
  title={Hamilton-Jacobi reachability: A brief overview and recent advances}, 
  year={2017},
  volume={},
  number={},
  pages={2242-2253},
  doi={10.1109/CDC.2017.8263977},
  ISSN={},
  month={Dec},}

@article{ChenHJReachability,
author = {Chen, Mo and Tomlin, Claire},
year = {2018},
month = {05},
pages = {333-358},
title = {Hamilton–Jacobi Reachability: Some Recent Theoretical Advances and Applications in Unmanned Airspace Management},
volume = {1},
journal = {Annual Review of Control, Robotics, and Autonomous Systems},
doi = {10.1146/annurev-control-060117-104941}
}

@INPROCEEDINGS{TabbaraDesignLatent,
  author={Tabbara, Ihab and Yang, Yuxuan and Hamzeh, Ahmad and Astafyev, Maxwell and Sibai, Hussein},
  booktitle={2026 IEEE International Conference on Robotics and Automation (ICRA)}, 
  title={Designing Latent Safety Filters using Pre-Trained Vision Models}, 
  year={2026},
  volume={},
  number={},
  pages={18315-18322},
  doi={10.1109/ICRA57385.2026.11695923},
  ISSN={},
  month={June},}

@InProceedings{seoUncertainty-awareLatentSF,
  title = 	 {Uncertainty-aware Latent Safety Filters for Avoiding Out-of-Distribution Failures},
  author =       {Seo, Junwon and Nakamura, Kensuke and Bajcsy, Andrea},
  booktitle = 	 {Proceedings of The 9th Conference on Robot Learning},
  pages = 	 {4442--4472},
  year = 	 {2025},
  editor = 	 {Lim, Joseph and Song, Shuran and Park, Hae-Won},
  volume = 	 {305},
  series = 	 {Proceedings of Machine Learning Research},
  month = 	 {27--30 Sep},
  publisher =    {PMLR},
  url = 	 {https://proceedings.mlr.press/v305/seo25a.html}
}

@InProceedings{ZendelCamera+Lidar_fuse,
    author    = {Zendel, Oliver and Huemer, Johannes and Murschitz, Markus and Dominguez, Gustavo Fernandez and Lobe, Amadeus},
    title     = {Joint Camera and LiDAR Risk Analysis},
    booktitle = {Proceedings of the IEEE/CVF Conference on Computer Vision and Pattern Recognition (CVPR) Workshops},
    month     = {June},
    year      = {2023},
    pages     = {88-97}
}

@ARTICLE{SylviaHJR_survey,
  author={Ganai, Milan and Gao, Sicun and Herbert, Sylvia L.},
  journal={IEEE Open Journal of Control Systems}, 
  title={Hamilton-Jacobi Reachability in Reinforcement Learning: A Survey}, 
  year={2024},
  volume={3},
  number={},
  pages={310-324},
  doi={10.1109/OJCSYS.2024.3449138},
  ISSN={2694-085X},
  month={},}

@INPROCEEDINGS{BajcsyReachability-Based_Framework,
  author={Bajcsy, Andrea and Bansal, Somil and Bronstein, Eli and Tolani, Varun and Tomlin, Claire J.},
  booktitle={2019 IEEE 58th Conference on Decision and Control (CDC)}, 
  title={An Efficient Reachability-Based Framework for Provably Safe Autonomous Navigation in Unknown Environments}, 
  year={2019},
  volume={},
  number={},
  pages={1758-1765},
  doi={10.1109/CDC40024.2019.9030133},
  ISSN={2576-2370},
  month={Dec},}

@ARTICLE{ChenFaSTrack,
  author={Chen, Mo and Herbert, Sylvia L. and Hu, Haimin and Pu, Ye and Fisac, Jaime Fernández and Bansal, Somil and Han, SooJean and Tomlin, Claire J.},
  journal={IEEE Transactions on Automatic Control}, 
  title={FaSTrack:A Modular Framework for Real-Time Motion Planning and Guaranteed Safe Tracking}, 
  year={2021},
  volume={66},
  number={12},
  pages={5861-5876},
  doi={10.1109/TAC.2021.3059838},
  ISSN={1558-2523},
  month={Dec},}

@ARTICLE{BorquezOnSafety,
  author={Borquez, Javier and Chakraborty, Kaustav and Wang, Hao and Bansal, Somil},
  journal={IEEE Transactions on Robotics}, 
  title={On Safety and Liveness Filtering Using Hamilton–Jacobi Reachability Analysis}, 
  year={2024},
  volume={40},
  number={},
  pages={4235-4251},
  doi={10.1109/TRO.2024.3454470},
  ISSN={1941-0468},
  month={},}

@article{BorquezHJ,
author = {Javier Borquez and Shuang Peng and Somil Bansal},
title ={Hamilton-Jacobi reachability for hybrid systems: Unified goal-driven control with safety guarantees},

journal = {The International Journal of Robotics Research},
volume = {0},
number = {0},
pages = {02783649261477777},
year = {2026},
doi = {10.1177/02783649261477777},

URL = { 
    
        https://doi.org/10.1177/02783649261477777
    
    

},
eprint = { 
    
        https://doi.org/10.1177/02783649261477777
    
    

}
}

@InProceedings{ToufighiDecisionboundary,
  title = 	 {Decision boundary learning for safe vision-based navigation via {H}amilton-{J}acobi reachability analysis and support vector machine},
  author =       {Toufighi, Tara and Bui, Minh and Shrestha, Rakesh and Chen, Mo},
  booktitle = 	 {Proceedings of the 6th Annual Learning for Dynamics \&amp; Control Conference},
  pages = 	 {440--452},
  year = 	 {2024},
  editor = 	 {Abate, Alessandro and Cannon, Mark and Margellos, Kostas and Papachristodoulou, Antonis},
  volume = 	 {242},
  series = 	 {Proceedings of Machine Learning Research},
  month = 	 {15--17 Jul},
  publisher =    {PMLR},
  url = 	 {https://proceedings.mlr.press/v242/toufighi24a.html}
}

@INPROCEEDINGS{NakamuraK, 
    AUTHOR    = {Kensuke Nakamura AND Lasse Peters AND Andrea Bajcsy}, 
    TITLE     = {{Generalizing Safety Beyond Collision-Avoidance via Latent-Space Reachability Analysis}}, 
    BOOKTITLE = {Proceedings of Robotics: Science and Systems}, 
    YEAR      = {2025}, 
    ADDRESS   = {LosAngeles, CA, USA}, 
    MONTH     = {June}, 
    DOI       = {10.15607/RSS.2025.XXI.113} 
}

@INPROCEEDINGS{Abdi,
  author={Abdi, Hossein and Raja, Golnaz and Ghabcheloo, Reza},
  booktitle={2023 IEEE International Conference on Robotics and Automation (ICRA)}, 
  title={Safe Control using Vision-based Control Barrier Function (V-CBF)}, 
  year={2023},
  volume={},
  number={},
  pages={782-788},
  doi={10.1109/ICRA48891.2023.10160805},
  ISSN={},
  month={May},}

@ARTICLE{Keyumarsi,
  author={Keyumarsi, Shaghayegh and Atman, Made Widhi Surya and Gusrialdi, Azwirman},
  journal={IEEE Robotics and Automation Letters}, 
  title={LiDAR-Based Online Control Barrier Function Synthesis for Safe Navigation in Unknown Environments}, 
  year={2024},
  volume={9},
  number={2},
  pages={1043-1050},
  doi={10.1109/LRA.2023.3339059},
  ISSN={2377-3766},
  month={Feb},}

@INPROCEEDINGS{DaSa,
  author={De Sa, Massimiliano and Kotaru, Prasanth and Sreenath, Koushil},
  booktitle={2024 IEEE International Conference on Robotics and Automation (ICRA)}, 
  title={Point Cloud-Based Control Barrier Function Regression for Safe and Efficient Vision-Based Control}, 
  year={2024},
  volume={},
  number={},
  pages={366-372},
  doi={10.1109/ICRA57147.2024.10610647},
  ISSN={},
  month={May},}

@article{Chang,
title = {Risk aware safe control with multi-modal sensing for dynamic obstacle avoidance},
journal = {Control Engineering Practice},
volume = {175},
pages = {107114},
year = {2026},
issn = {0967-0661},
doi = {https://doi.org/10.1016/j.conengprac.2026.107114},
url = {https://www.sciencedirect.com/science/article/pii/S0967066126003576},
author = {Pei Yu Chang and Qizhe Xu and Vishnu Renganathan and Qadeer Ahmed}
}

@ARTICLE{Gravina,
    
AUTHOR={Gravina, Giovanbattista  and D’Orazio, Francesco  and Cipriano, Michele  and Belvedere, Tommaso  and Oriolo, Giuseppe },
           
TITLE={Crowd navigation in a multi-room environment: a model predictive control framework for mobile robots},
          
JOURNAL={Frontiers in Robotics and AI},
          
VOLUME={Volume 13 - 2026},
  
YEAR={2026},
  
URL={https://www.frontiersin.org/journals/robotics-and-ai/articles/10.3389/frobt.2026.1812386},
  
DOI={10.3389/frobt.2026.1812386},
  
ISSN={2296-9144}}

@inproceedings{Dino,
  title={Emerging Properties in Self-Supervised Vision Transformers},
  author={Caron, Mathilde and Touvron, Hugo and Misra, Ishan and J{\'e}gou, Herv{\'e} and Mairal, Julien and Bojanowski, Piotr and Joulin, Armand},
  booktitle={Proceedings of the IEEE/CVF International Conference on Computer Vision (ICCV)},
  pages={9650--9660},
  year={2021}
}

@inproceedings{SpUNet,
  title={4D Spatio-Temporal ConvNets: Minkowski Convolutional Neural Networks},
  author={Choy, Christopher and Gwak, JunYoung and Savarese, Silvio},
  booktitle={Proceedings of the IEEE/CVF Conference on Computer Vision and Pattern Recognition (CVPR)},
  pages={3075--3084},
  year={2019}
}

@inproceedings{siddiqui2024assetgen,
  title={Meta 3D AssetGen: Text-to-Mesh Generation with High-Quality Geometry, Texture, and PBR Materials},
  author={Siddiqui, Yawar and Monnier, Tom and Kokkinos, Filippos and Kariya, Mahendra and Kleiman, Yanir and Garreau, Emilien and Gafni, Oran and Neverova, Natalia and Vedaldi, Andrea and Shapovalov, Roman and Novotny, David},
  booktitle={Advances in Neural Information Processing Systems},
  volume={37},
  year={2024}
}

@article{sam3d2025,
  title={SAM 3D: 3Dfy Anything in Images},
  author={{SAM 3D Team} and Chen, Xingyu and Chu, Fu-Jen and Gleize, Pierre and others},
  journal={arXiv preprint arXiv:2511.16624},
  year={2025},
  doi={10.48550/arXiv.2511.16624},
  url={https://arxiv.org/abs/2511.16624}
}

@inproceedings{schonberger2016structure,
  title={Structure-from-Motion Revisited},
  author={Sch{\"o}nberger, Johannes L. and Frahm, Jan-Michael},
  booktitle={Proceedings of the IEEE Conference on Computer Vision and Pattern Recognition},
  pages={4104--4113},
  year={2016},
  doi={10.1109/CVPR.2016.445}
}

@article{kerbl2023gaussian,
  title={3D Gaussian Splatting for Real-Time Radiance Field Rendering},
  author={Kerbl, Bernhard and Kopanas, Georgios and Leimk{\"u}hler, Thomas and Drettakis, George},
  journal={ACM Transactions on Graphics},
  volume={42},
  number={4},
  pages={139:1--139:14},
  year={2023},
  doi={10.1145/3592433}
}

@inproceedings{huang2024gaussian,
  title={2D Gaussian Splatting for Geometrically Accurate Radiance Fields},
  author={Huang, Binbin and Yu, Zehao and Chen, Anpei and Geiger, Andreas and Gao, Shenghua},
  booktitle={ACM SIGGRAPH 2024 Conference Papers},
  articleno={32},
  year={2024},
  doi={10.1145/3641519.3657428}
}

@inproceedings{wu20253dgut,
  title={3DGUT: Enabling Distorted Cameras and Secondary Rays in Gaussian Splatting},
  author={Wu, Qi and Martinez Esturo, Janick and Mirzaei, Ashkan and Moenne-Loccoz, Nicolas and Gojcic, Zan},
  booktitle={Proceedings of the IEEE/CVF Conference on Computer Vision and Pattern Recognition},
  pages={26036--26046},
  year={2025}
}

@manual{unitreeG1Manual,
  author = {{Unitree Robotics}},
  title = {Unitree G1 User Manual},
  year = {2024},
  url = {https://marketing.unitree.com/article/en/G1/User_Manual.html},
  note = {Accessed: 2026-11-11}
}

@misc{livoxMid360,
    author = {{Livox}},
    title = {MID-360},
    howpublished = {\url{https://www.livoxtech.com/mid-360}},
    note = {Accessed: November 11, 2026}
}

@misc{realsenseD435i,
    author = {{Intel RealSense}},
    title = {Intel RealSense Depth Camera D435i},
    howpublished = {\url{https://www.intelrealsense.com/depth-camera-d435i/}},
    note = {Accessed: November 11, 2026}
}
\end{spacing}



\end{document}